\documentclass[review]{elsarticle}

\usepackage{lineno}
\modulolinenumbers[5]

\usepackage{amsmath,amssymb,amsfonts,bm}
\usepackage{caption}
\usepackage{multirow}
\usepackage{subfigure}
\usepackage{pifont}
\usepackage{float}
\usepackage{makecell}
\usepackage{booktabs}
\usepackage{multirow}
\usepackage{colortbl}
\usepackage[dvipsnames]{xcolor}
\usepackage{bbding}
\usepackage[ruled]{algorithm2e}

\usepackage[colorlinks, linkcolor=blue, anchorcolor=blue, citecolor=blue]{hyperref}
\usepackage{geometry}
\journal{Journal of Elsevier}

\begin{document}

\begin{frontmatter}

\title{Differential Evolution based Dual Adversarial Camouflage: Fooling Human Eyes and Object Detectors}

\author[mysecondaryaddress]{Jialiang Sun}

\author[mysecondaryaddress]{Wen Yao\corref{mycorrespondingauthor}}
\ead{wendy0782@126.com}
\author[mysecondaryaddress]{Tingsong Jiang \corref{mycorrespondingauthor}}
\cortext[mycorrespondingauthor]{Corresponding author}
\ead{tingsong@pku.edu.cn}

\author[mymainaddress]{Donghua Wang}
\author[mysecondaryaddress]{Xiaoqian Chen}

\address[mysecondaryaddress]{Defense Innovation Institute, Chinese Academy of Military Science, No. 53, Fengtai East Street, Beijing 100071, China}
\address[mymainaddress]{College of Computer Science and Technology, Zhejiang University}

\begin{abstract}
Deep neural network-based object detectors are vulnerable to adversarial examples. Among existing works to fool object detectors, the camouflage-based method is more often adopted due to its adaptation to multi-view scenarios and non-planar objects. However, most of them can still be easily observed by human eyes, which limits their application in the real world. To fool human eyes and object detectors simultaneously, we propose a differential evolution based dual adversarial camouflage method. Specifically, we try to obtain the camouflage texture by the two-stage training, which can be wrapped over the surface of the object. In the first stage, we optimize the global texture to minimize the discrepancy between the rendered object and the scene background, making human eyes difficult to distinguish. In the second stage, we design three loss functions to optimize the local texture, which is selected from the global texture, making object detectors ineffective. In addition, we introduce the differential evolution algorithm to search for the near-optimal areas of the object to attack, improving the adversarial performance under certain attack area limitations. Experimental results show that our proposed method can obtain a good trade-off between fooling human eyes and object detectors under multiple specific scenes and objects.
\end{abstract}

\end{frontmatter}

\section{Introduction}

Deep neural network (DNN) have made great progresses in many computer vision tasks such as the image classification \cite{Remote2022,wang2017residual,lu2007survey} and the object detection \cite{QiangInstant,zhao2019object,szegedy2013deep}. However, DNN models proved to be vulnerable to adversarial examples (AEs) that are intentionally crafted with the perturbations, which can make DNN models output wrong results \cite{goodfellow2014explaining,wang2021dualdaac,pgd2018}. The phenomenon of AEs causes a significant threat to the real-world application of DNN models such as self-driving or facial recognition systems. To explore the robustness of DNN models and guide their further practical application, adversarial attack algorithms have been extensively exploited for the majority of tasks, which aim to study how to generate AEs to fool DNN models.

According to the magnitude of the perturbation, the adversarial attack can be categorized into $l_{p}$ norm based adversarial attack and unrestricted adversarial attack. In $l_{p}$ norm based adversarial attack, the $l_{p}$ norm of the generated perturbation can not exceed a predefined threshold, which can ensure that the perturbations are imperceptible \cite{li2022approximated,li2023adaptive}. Currently, extensive works about the $l_{p}$ norm based adversarial attack are developed such as fast gradient sign method (FGSM) \cite{goodfellow2014explaining}, projection gradient descent (PGD) \cite{pgd2018} attack and Carlini $\&$ Wagner (C$\&$W) attack \cite{cw2017}. However, the AEs generated by the $l_{p}$ norm based attacks can not be printed in the real physical-world, which limits the further application in complicated tasks such as object detection. Thus in the physical-world attacks, unrestricted adversarial attacks are more adopted, which do not limit the magnitude or area of the generated perturbations. However, unrestricted adversarial attacks will cause the
conspicuous appearance. Hence, how to balance the adversarial performance and naturalness towards human eyes is a challenging problem.

To realize the balance between adversarial performance and naturalness in unrestricted perturbations, some works have been devoted to image classification and object detection tasks. In the classification task, some works first tried to change the color or lightness of each image \cite{hosseini2018semantic,laidlaw2019functional}. Another direction to generate unrestricted AEs is by simulating the natural weather, such as fog and snow \cite{hendrycks2019benchmarking}. To approach the physical-world attack, Duan et al. \cite{duan2020adversarial} further combined the style transfer technology with the adversarial attack to make the generated AEs maintain the specific style. In the object detection task, existing methods can be divided into patch-based and camouflage-based. The patch-based method tried to optimize a adversarial patch pasted on the object, which confines
the noise to a small and localized patch without perturbation
constraint. A patch is often stuck to the planar objects like stop sign \cite{eykholt2018robust}, placed in the front of objects like object person \cite{thys2019fooling} or placed on the background of images \cite{liu2018dpatch}. In contrast, the camouflage-based method is implemented by modifying the target object itself and is more chanllenging due to the non-planarity of 3D objects. There are two ways to obtain the camouflage. One way is to optimize the despired pattern. Similar work is like dual attention suppression (DAS) attack proposed by Wang et al. \cite{wang2021dual}, which constrains the perturbations into the shape of a smile to obtain the camouflage that can be pasted on the surface of cars. The other way is to optimize the camouflage to paint on the full surface of the object. For instance, Huang et al. \cite{huang2020universal} proposed universal physical camouflage (UPC) towards person object detectors, which looks like the dogs. Wang et al. \cite{wang2022fca} proposed the full camouflage attack (FCA), which optimizes a jungle camouflage that can be wrapped on the full surface of cars to fool detectors in city roads dataset. In general, the camouflage-based method is more practical in the real world under multi-view scenarios.

Though these works about balancing the adversarial performance and naturalness of adversarial examples have been exploited, it is still challenging to generate the AEs that can fool human eyes and object detectors simultaneously. First, since the AEs of most of the works in classification tasks are stationary and designed for the single image, they can not be transferred to the object detection task directly. For instance, the adversarial camouflage proposed by Dual et al. \cite{duan2020adversarial} has a natural visual effect in the classification task, which can matain the specific style. But it is not suitable for moving objects like soldiers and cars, because it is
difficult to select the suitable image style to match the constantchanging
backgrounds. Second, existing camouflages towards object detectors are usually required to ensure their own naturalness, like the simile face, which also can not be adaptive to the changing environments due to the problem that AEs are easily observed by human eyes. 

\begin{figure}[h]
	\centering
	\includegraphics[scale=0.35]{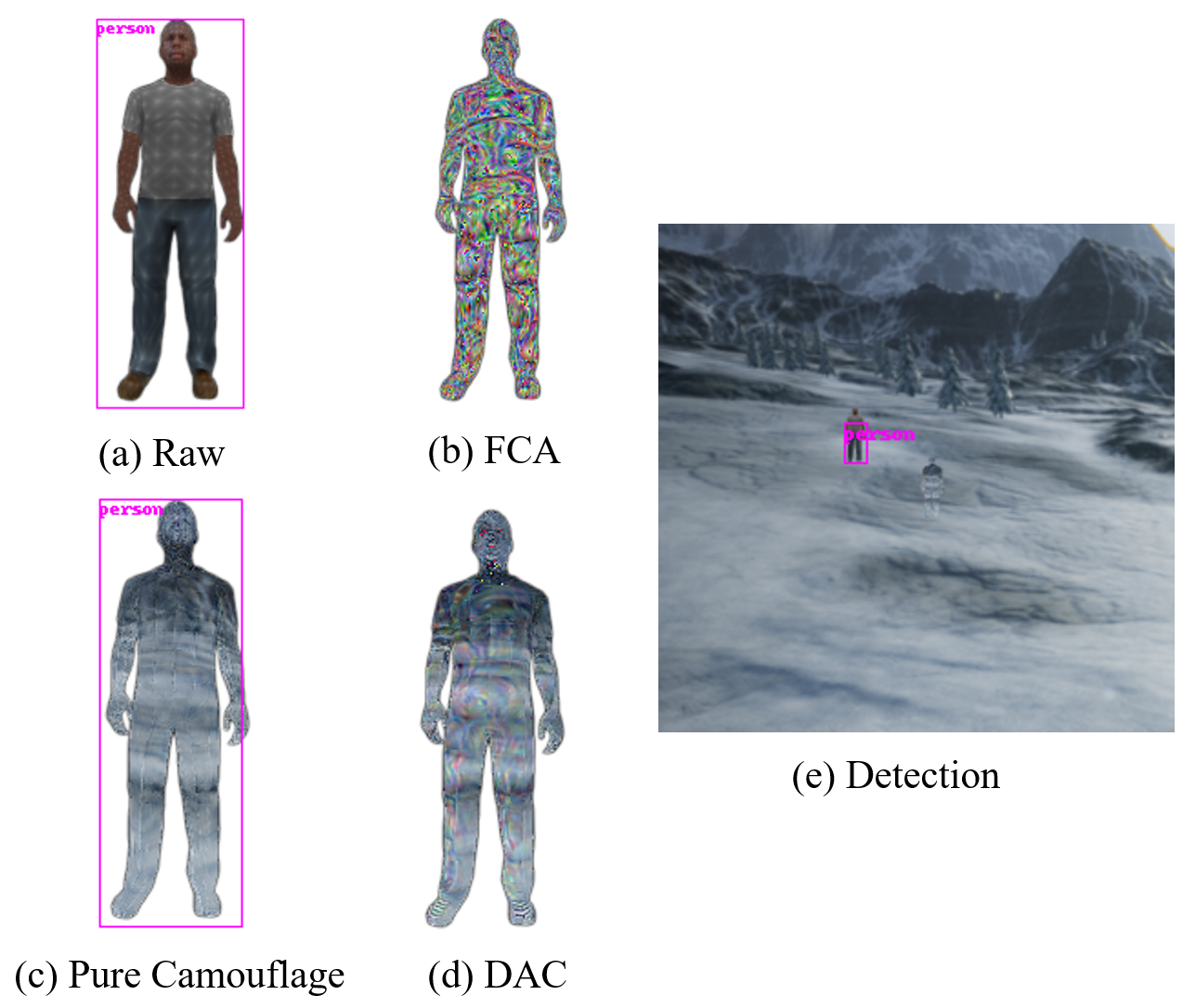}
	\caption{(a) is the person without camouflage. (b) is the camouflaged person by FCA. (c) is the camouflaged person using the pure color constraint. (d) is the camouflage produced by DAC. (e) shows the final detection performance of (a) and (d) in the specific scene.}
	
	\label{final}
\end{figure}

To bridge the gap, given the specific scene, we propose an end-to-end differential evolution based dual adversarial camouflage (DE$\_$DAC) pipeline, which can fool human eyes and object detection simultaneously. Specifically, treating the generation of camouflage as the optimization of the texture that can be painted on the surface of objects, we design a two stage training method based on the neural render. In the first stage, given the training dataset, we select several background images as scene images. We train the global texture to make the rendered image close to the scene images. After the first stage ends, the rendered object can be well integrated with the specific scene, which makes the human eyes uneasy to distinguish. In the second stage, we select the local texture from the optimized global texture, the number of which can be set flexibly. We design three loss functions to optimize the local texture. The adversarial loss is to enhance the adversarial performance. The smooth loss can be utilized to ensure the printness of the final texture, while the color loss is developed to minimize the MSE between the local texture and the global texture. The final texture is the combination of the global texture and local texture. The whole process is denoted as dual adversarial camouflage (DAC). To further improve the performance of our proposed method, under the same number of faces of the model that can represent the position of local texture, we introduce the differential evolution (DE) algorithm to search for the near-optimal attack area of the second stage. In addition, motivated by the recently developed robot that can change color in real-time according to the background \cite{kim2021biomimetic}, we further study the performance of adaptive DE$\_$DAC (A-DE$\_$DAC). In A-DE$\_$DAC, we generate the adaptive camouflage according to each scene image in the first stage, realizing that the generated camouflage can change with the environment. One example of our proposed method is presented in Figure \ref{final}. Figure \ref{final} (c) is the optimized camouflage in the first stage to fool human eyes. Figure \ref{final} (d) is the final camouflage when adversarial performance is considered in the second optimization stage. Figure \ref{final} (e) presents the detection comparison of the clean image and the optimized camouflage by us. From Figure \ref{final} (e), we can see that our generated camouflage can fool object detectors and human eyes well.

Our contributions can be summarized as follows:

\begin{itemize}
	\item We propose DE based dual adversarial camouflage to generate the camouflage that obtains a good trade-off between fooling the human eyes and object detectors, which can also be easily combined with different scenes and objects.
	\item We introduce the two-stage training strategy to obtain the camouflage. In the first stage, we optimize the global texture to make the rendered object look close to the specific scenes. In the second stage, we optimize the local texture to fool the object detectors.
	\item We introduce the DE algorithm to search for the near-optimal area to attack, improving the attack success rate with the limitation of areas.
	\item We further study the performance of adaptive DE$\_$DAC, where the generated camouflage can change with the environment. Under the same selected local texture, compared with DE$\_$DAC, adaptive DE$\_$DAC can further improve the camouflage performance.

\end{itemize}

The remainder of this paper is organized as follows. Section \ref{sec2} introduces the preliminary knowledge about adversarial attack, natural adversarial examples and neural render. Section \ref{sec3} introduces our proposed method. Section \ref{sec4} elaborates the experimental settings and results. Section \ref{sec5} makes the conclusions and discussions.

\section{Background}\label{sec2}

\subsection{Natural adversarial example}

\begin{figure*}[htbp]
	\centering
	\subfigure[LAP]{
		\includegraphics[height=0.18\linewidth]{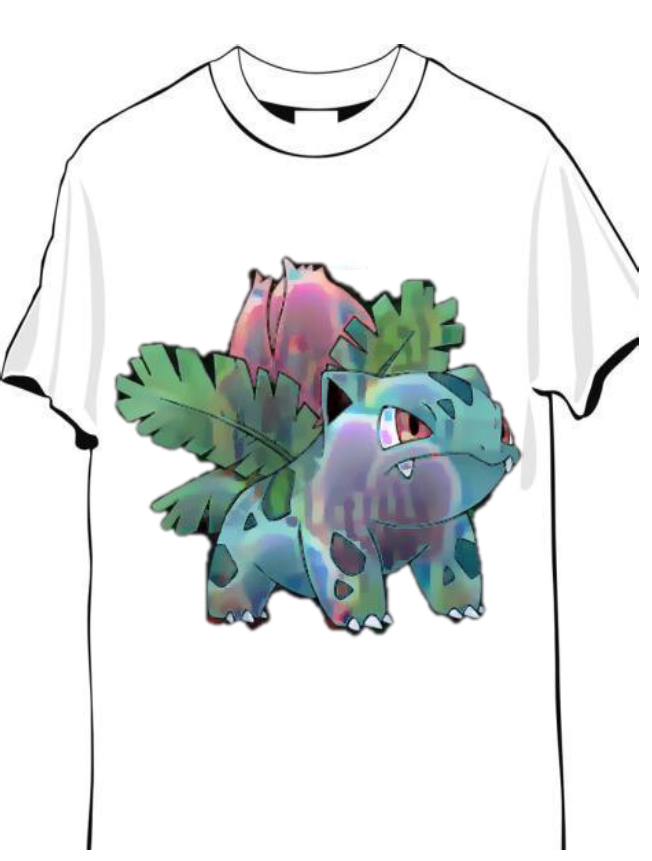}
	}
	\subfigure[AdvCam]{
		\includegraphics[height=0.18\linewidth]{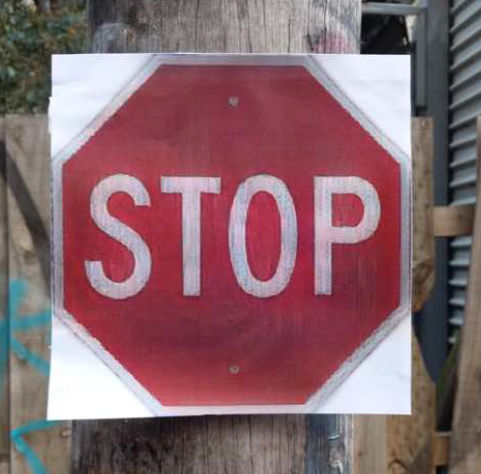}
	}
	\subfigure[UPC]{
		\includegraphics[height=0.18\linewidth]{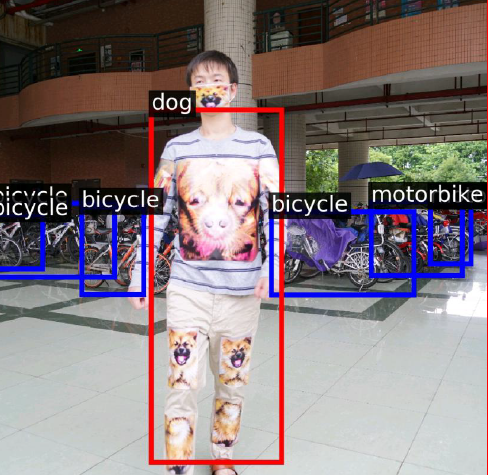}
	}
	\subfigure[CAMOU]{
		\includegraphics[height=0.18\linewidth]{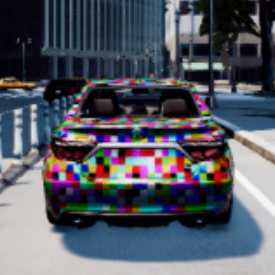}
	}
	\subfigure[DAS]{
		\includegraphics[height=0.18\linewidth]{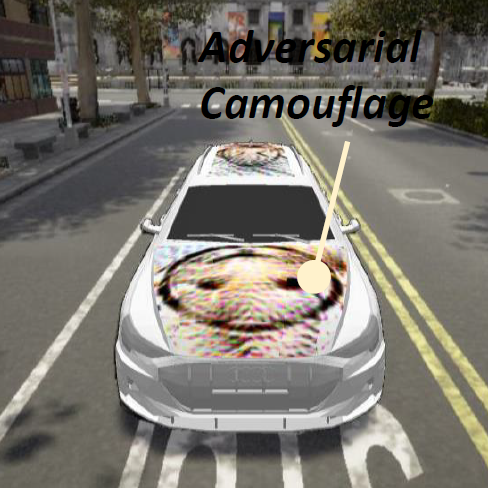}
	}	
	\caption{The visualization of the optimized camouflage using different methods.}
	\label{relatedwork}
\end{figure*}

Due to the unlimitation of the magnitude in perturbation, how to realize the naturalness of AEs towards human eyes remains a challenging problem. Many works focused on the goal that the optimized AEs possess their own naturalness. Tan et al. \cite{tan2021legitimate} proposed the legitimate adversarial patches (LAP) to fool object detectors, which constrains the edge and texture close to the pre-selected natural image in the process of training the patch. Duan et al. \cite{duan2020adversarial} proposed AdvCam, which combines the style loss, smooth loss and adversarial loss together to train the adversarial example, which is verified in the classification task. Huang et al. \cite{huang2020universal} propose universal physical camouflage (UPC) towards object detectors. The final AEs look like the dogs. Wang et al. \cite{wang2021dual} proposed a dual attention suppression attack, which optimizes a smile texture pasted on the surface of the cars. Zhang et al. \cite{2018CAMOU} proposed CAMOU, which learns a 3D camouflage by repeating the images on the surface of objects. Some of the works mentioned above are visualized in Figure \ref{relatedwork}. In general, most of the existing works focus on the methodology of constraining the optimized camouflage toward the specific shape or texture. Though they can obtain the trade-off between the adversarial performance and the naturalness of the optimized camouflage itself, the final adversarial example can still be easily observed by human eyes. 

\subsection{Neural render}

\begin{figure*}[h]
	\centering
	\includegraphics[scale=0.71]{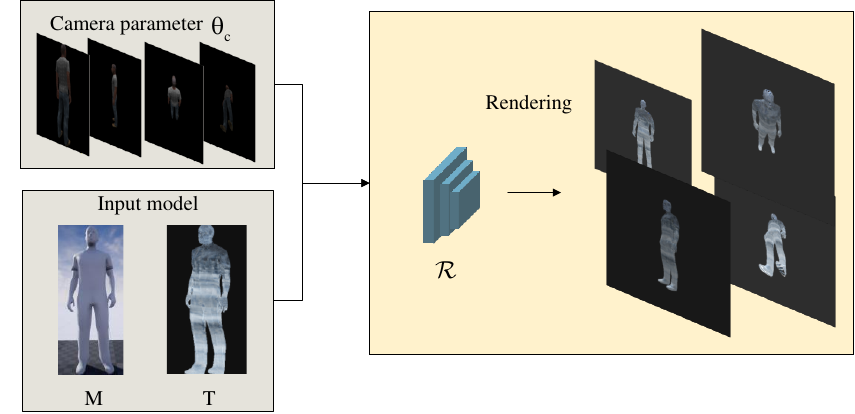}
	\caption{The process of generating the rendered person image using neural render.}
	
	\label{nr}
\end{figure*}

The goal of the traditional renderer is to realize the 2D to 3D transformation. One of the applications is to wrap the texture image to the 3D model, which is then rendered to the 2D image. To make the process of rendering differentiable, Kato et al. \cite{kato2018neural} proposed the neural render, which utilized the approximate gradient for rasterization to integrate the rendering into neural networks. By initializing the texture with different camera parameters, such as the rotation and location, one could render the 3D object model
consisting of mesh and texture under different view angles. Due to the ability to approach the physical conditions, neural render has been 
utilized in the CARLA \cite{dosovitskiy2017carla} simulator to render the adversarial
patch onto the 3D object.
Xiao et al. \cite{xiao2019meshadv} also used the neural renderer to modify the
shape and texture of 3D objects. Following Wang et al. \cite{wang2021dual}, we utilize the neural renderer to paint our adversarial
camouflage onto the surface of 3D object. Figure \ref{nr} presents an example of the rendered person image using neural render $\mathcal{R}$. Given the input model consisting of 3D model $\textbf{M}$ and its texture $\textbf{T}$, by controlling the camera parameters $\theta_{c}$, we can obtain different rendered person images with multi-views. Each rendered person image just includes the person and does not have the background.

\section{Method}\label{sec3}
In this section, the proposed DE based dual adversarial camouflage is introduced in detail. In section \ref{sec31}, the overview of dual adversarial camouflage generation framework is presented. In section \ref{sec311} and section \ref{sec312}, the specific strategies of two-stage in DAC are elaborated respectively. In section \ref{sec32}, the DE based search for near-optimal attack areas is described. In section \ref{sec33}, depending on whether the camouflage can be adapted to the scene, A-DE$\_$DAC is further introduced.
\subsection{Overview of DAC}\label{sec31}

\begin{figure*}[h]
	\centering
	\includegraphics[scale=0.8]{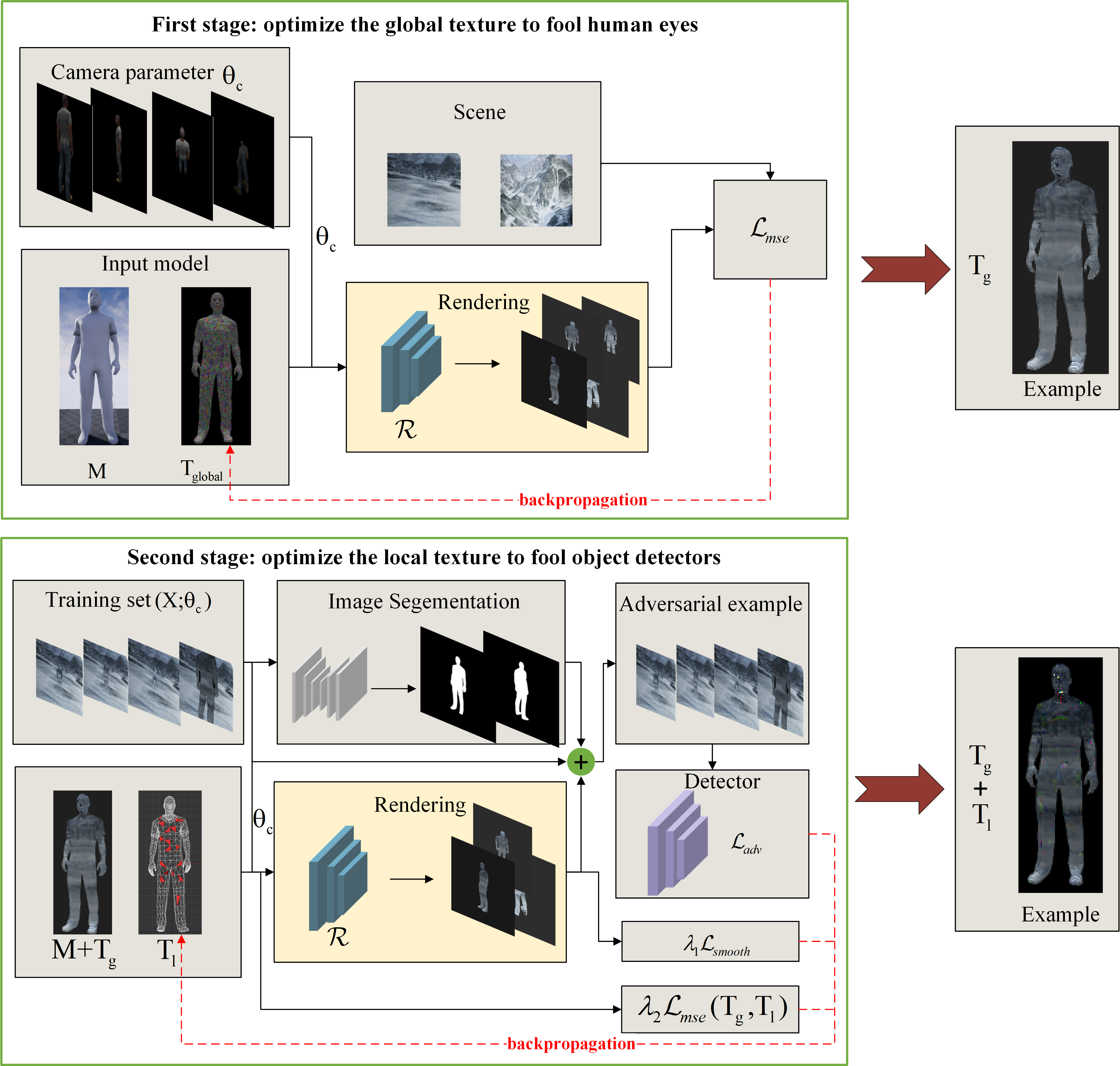}
	\caption{The overview of the proposed DAC. The whole process can be divided into two stages. In the fist stage, 	we optimize the global texture to obtain the camouflage to fool human eyes. Under different camera parameters, the MSE between the rendered image and the scene image is as the loss function. In the second stage, we optimize the local texture to fool object detector.}
	
	\label{dadd}
\end{figure*}

Our goal is to generate the camouflage to fool the human eyes and object detectors simultaneously. Our proposed DAC method includes two stages. In the first stage, we minimize the mean square error (MSE) between the scene images and the rendered images to obtain the texture, which could fool human eyes easily. In the second stage, on the given training dataset, we optimize the local texture to enhance the performance of adversarial attacks. The overview of our proposed DAC method is presented in Figure \ref{dadd}.

\subsubsection{First stage: generate the global texture to fool human eyes}\label{sec311}
In the first stage, the goal of the optimization process is to obtain the camouflage that fools human eyes under the specific scene. Given some scene images selected from the specific scene, we aim to minimize the gap between the rendered person images and the scene images. Thus the loss function of optimizing the global texture is expressed as follows:
\begin{equation}
\begin{array}{l}
\mathcal{L}_{first} = \frac{\sum \limits_{i=1} \sum \limits_{j=1} {\text{MSE}(\mathbf{O}_{i},\textbf{I}_{j})}}{n}
\end{array}
\label{natural}
\end{equation}
where $\textbf{I}_{j}$ denotes the $j_{th}$ scene image selected from the training dataset. $\mathbf{O}_{i}$ stands for the $i_{th}$ rendered image using the texture $\textbf{T}_{g}$. $n$ denotes the total number of final images that consist of rendered person images and scene images. We can obtain the trained $\textbf{T}_{g}$ by minimizing the MSE loss between the rendered image and scene image. In this stage, the camera parameters are manually-controlled to ensure that the generated camouflage is natural enough. For instance, near camera distance and  360-degree rotation perspective can be better for training the final texture. Detailed operations can also be seen in neural render introduced by Kennedym et al. \cite{Kennedym2002}. 

\subsubsection{Second stage:generate the local texture to fool object detectors}\label{sec312}

In the second stage, the goal of the optimization process is to obtain the camouflage that fools the object detector with high naturalness. To realize the purpose, we retrain the selected part in the global texture by the first stage. The 3D model is rendered by multiple polyhedrons, which can be described as faces. Assuming the number of faces of the model is $n_{m}$, the continuous sequence of the face [1,2,...,$n_{m}$] can stand for the full surface. If we need to select the part of the surface such as $n_{f}$ faces, we can generate the sequence consisting of $n_{f}$ non-repeatable values from [1,$n_{m}$]. The optimized local texture is denoted as $\textbf{T}_{l}$. By introducing the mask $\textbf{M}_{l}$ consisting of zero and one determined by the faces, we could merely optimize the wanted positions. The one in $\textbf{M}_{l}$ stands for the positions of texture $\textbf{T}_{l}$, and other parameters could be obtained from the   $\textbf{T}_{g}$. Thus, the final texture $\textbf{T}_{adv}$ could be obtained as follows.
\begin{equation}
\begin{array}{l}
\textbf{T}_{adv} = \textbf{T}_{g} \cdot (1-$\textbf{M}$_{l}) + $\textbf{M}$_{l} \cdot (\textbf{T}_{l})
\end{array}
\label{finalT}
\end{equation}

After obtaining $\textbf{T}_{l}$, we can obtain the final texture $\textbf{T}_{adv}$ according to Eq. \ref{finalT}. Then we can obtain the rendered person image $\mathbf{O}$. 

\begin{equation}
\mathbf{O} = \mathcal{R}((\textbf{M},\textbf{T}_{adv});\theta_{c})
\end{equation}

The segment network is utilized to obtain the mask $m$ of $\mathbf{O}$, where one stands for the object while zeros represents the background. The final AEs can be composed of  $\mathbf{O}$ and $\mathbf{I}$ by $m$, as presented in Eq. \ref{transfor}.

\begin{equation}
\mathbf{I}_{a d v}=\Phi(\mathbf{O})=m \cdot \mathbf{O}+(1-m) \cdot \mathbf{I}
\label{transfor}
\end{equation}

To obtain the near-optimal $\textbf{T}_{l}$, we design the combination of three loss functions, including the adversarial loss $\mathcal{L}_{adv}$, color loss $\mathcal{L}_{color}$ and smooth loss $\mathcal{L}_{smooth}$. These three loss functions are described as follows.

\begin{itemize}
	\item Adversarial Loss
\end{itemize}
In our work, the classical Yolo-V3 object detector is adopted as the white-box target model. The adversarial loss as follow. 
\begin{equation}
\begin{array}{l}
\mathcal{L}_{adv} = \mathcal{L}_{obj}(\mathbf{I}_{a d v})
\end{array}
\label{attack}
\end{equation}
where $\mathcal{L}_{obj}$ represents the confidence score
whether the detection box contains an object. For other object detectors, the specific $\mathcal{L}_{adv}$ could be designed adaptively.

\begin{itemize}
	\item Color Loss
\end{itemize}
To ensure the naturalness of the generated camouflage by local texture $\textbf{T}_{l}$, in the second stage, the color loss is introduced as a constraint. We aim to make the discrepancy between the local texture and the obtained global texture $\textbf{T}_{g}$ close, named $\mathcal{L}_{color}$, which is expressed as follows.
\begin{equation}
\begin{array}{l}
\mathcal{L}_{color} = \sum [(\textbf{T}_{g} - \textbf{T}_{l}) \cdot $\textbf{M}$_{l}]^{2}
\end{array}
\label{color2}
\end{equation}

\begin{itemize}
	\item Smooth Loss
\end{itemize}
To ensure the printness of the generated
adversarial camouflage, we follow Wang et al. \cite{wang2022fca} to
utilize the smooth loss that introduced by Mahendran et al. \cite{mahendran2015understanding}, which can reduce the inconsistent among adjacent
pixels. For a rendered person image painted with adversarial
camouflage, the calculation of smooth loss can be
written as  
\begin{equation}
\begin{array}{l}
\mathcal{L}_{s m o o t h}=\sum_{i, j}\left(x_{i, j}-x_{i+1, j}\right)^{2}+\left(x_{i, j}-x_{i, j+1}\right)^{2}
\end{array}
\label{smmoth}
\end{equation}
where $x_{i, j}$ is the pixel value of $\textbf{I}_{adv}$ at coordinate $(i,j)$.

Combining three above loss functions, we could obtain the local texture:
\begin{equation}
\begin{array}{l}
\mathcal{L}_{second}=\mathcal{L}_{adv} + \lambda_{1} \mathcal{L}_{color} + \lambda_{2}\mathcal{L}_{s m o o t h}
\end{array}
\label{total}
\end{equation}
where $\lambda_{1}$, $\lambda_{2}$ and $\lambda_{3}$ are the coefficients to control the weight of loss values. Larger $\lambda_{2}$ means higher attack success rate, while smaller one would make the optimization process focus on the naturalness. To sum up, the whole process of generating the camouflage is presented in the following DAC algorithm.

\begin{algorithm}[!ht]	\label{alg1}
	\caption{Dual Adversarial Camouflage (DAC)}
	\LinesNumbered
	\KwIn{ 
		Training set ($\textbf{X},\theta_{c}$), 3D model ($\textbf{M,T}$),neural render $\mathcal{R}$, object detector $\mathcal{F}$, segmentation network $\mathcal{U}$, the mask $\textbf{M}_{l}$ determined by the selected faces of model.
	}
	\KwOut{Adversarial texture $\textbf{T}_{adv}$.}
	
	Initialize the $\textbf{T}_{g}$ with the random noise; \\ 
	\For{the maximum epoch of the first stage}{
		Select the mini-batch sample from the training set ($\textbf{X},\theta_{c}$);\\
		Extract the background of the selected sample using $\mathcal{U}$;\\
		Generate the rendered image $O$ $\leftarrow$  $\mathcal{R}((\textbf{M},\textbf{T}_{g});\theta_{c})$;\\
		Calculate the loss function of the first stage $\mathcal{L}_{first}$ by Eq.~\ref{natural};\\
		Update the $\textbf{T}_{g}$ by gradient back-propagation;\\
	}
	Initialize the $\textbf{T}_{l}$ with the random noise;\\
	\For{the maximum epoch of the second stage}{
		Select the mini-batch sample from the training set ($\textbf{X},\theta_{c}$);\\
		Extract the mask of the render image $m$ $\leftarrow$  $\mathcal{U}(\textbf{X})$;\\
		Obtain the final texture $\textbf{T}_{adv} $ $\leftarrow$$ \textbf{T}_{g}\cdot (1-$\textbf{M}$_{l}) + $\textbf{M}$_{l} \cdot (\textbf{T}_{l})$;\\
		Generate the rendered image $\mathbf{O}$ $\leftarrow$ $\mathcal{R}((\textbf{M},\textbf{T}_{adv});\theta_{c})$;\\
		Extract the background $\mathbf{I}$ of the selected sample using $\mathcal{U}$;\\
		Obtain the image composed of the reder image and background $\mathbf{I}_{a d v}$ $\leftarrow$ $m \cdot \mathbf{O}+(1-m) \cdot \mathbf{I}$;\\
		Calculate the $\mathcal{L}_{adv}$ using $\mathbf{I}_{a d v}$ by Eq.~\ref{attack} for the object detector $\mathcal{F}$;\\
		Calculate the $\mathcal{L}_{color}$ using $\textbf{T}_{g}$ and $\textbf{T}_{adv}$ by Eq.~\ref{color2};\\
		Calculate the $\mathcal{L}_{smooth}$ using $\mathbf{O}$ by Eq.~\ref{smmoth};\\
		Obtain the total loss $\mathcal{L}_{second}$ by Eq.~\ref{total};\\
		Update the $\textbf{T}_{l}$ by gradient back-propagation;\\
	}
	
\end{algorithm}

\subsection{DE based search for near-optimal attack area}\label{sec32}

In this section, we focus on improving the DAC method by DE based search for the near-optimal attack area when the number of faces of the models is settled. The mathematical model of searching for the near optimal attack area can be described as follows.

\begin{equation}
\begin{array}{ll}
\min\limits_{x}  & f(\textbf{x}) \\
\text { s.t. } 
&1\le {\textbf{x}_{i}}\le {{n}_{m}},\forall i\in \{1,2,\cdots ,{{n}_{f}}\}\\
& {\textbf{x}_{i}}\ne {{x}_{j}},\forall i,j\in \{1,2,\cdots ,{{n}_{f}}\} \\
\end{array}
\label{eqsingle}
\end{equation}
where $\textbf{x}$ denotes the selected face of the model, ${n}_{f}$ is the preset maximum number of selected faces, ${n}_{m}$ stands for the maximum number of faces the model with full surface, $f(\textbf{x})$ is the p@0.5 value of $\textbf{x}$ solved by the proposed DAC method in the dataset.

\begin{algorithm}[!ht]\label{alg2}
	\caption{DE based Dual Adversarial Camouflage (DE$\_$DAC)}
	\LinesNumbered
	\KwIn{The population size $n$,  the number of faces $n_{f}$, the maximum number of faces of the model $n_{m}$, the maximum iteration number $T$, crossover rate $r_{c}$, mutation rate $r_{m}$.
	}
	\KwOut{ The optimal face $\textbf{x}_{best}$.}
	Iteration counter number $i=1$;\\
	Population Initialization $\textbf{P}$ $\leftarrow$ generate $n$ individuals randomly, each of which is one sequence consisting of $n_{f}$ non-repeatable values from [1,$n_{m}$]; \\ 
	$fitness$($\textbf{P}$) $\leftarrow$  Evaluate the \textbf{p@0.5} value of each individual in population $\textbf{P}$ \textbf{using DAC algorithm};\\
	\For{$i<=T$}{
		$\textbf{P}_{m}$$\leftarrow$ Population mutation based on $\textbf{P}$;\\
		$\textbf{P}_{c}$  $\leftarrow$ Population crossover based on $\textbf{P}_{m}$;\\
		//  \textbf{Population selection}\\
		Individual counter number $j = 1$;\\
		\For{$j <= n$}{
			\If {fitness $(\textbf{P}_{c})[j]$  $<$ fitness$(\textbf{P})[j]$}
			{$\textbf{P}[j]$ = $\textbf{P}_{c}[j]$;}
			$j=j+1$;
		}
		
		Obtain the best individual $\textbf{x}_{best}$ $\leftarrow$ $argmin$($fitness$(\textbf{P}));\\
		Update $\textbf{P}$ by removing the duplicate numbers; \\
		$i=i+1$;\\
	}
\end{algorithm}

Then we introduce the DE algorithm to search for the near-optimal attack area in the second stage of DAC. The whole framework is denoted as DE$\_$DAC. DE is one classical gradient-free evolutionary algorithm to solve the black-box optimization problem, which is composed of population initialization, mutation, crossover, and selection operations. To realize the purpose of searching for the near-optimal attack area, as the description in section \ref{sec33}, we generate the population consisting of different individuals. Each individual is a sequence, consisting of $n_{f}$ non-repeatable values from [1,$n_{m}$], where $n_{f}$ is the pre-settled number of faces. During the optimization process, the p@0.5 value by DAC is adopted as the evaluation metric when the individual is determined. By the subsequent operations, including population crossover, mutation, and selection, the near-optimal sequence can be obtained, which corresponds to the near-optimal attack area in DE$\_$DAC. In this process, we need to remove the duplicate numbers in the searched sequence and randomly generate that again at each iteration. The pseudo-code of DE$\_$DAC is shown in the DE$\_$DAC algorithm.

\subsection{Adaptive DE$\_$DAC}\label{sec33}

Based on the formulations of above mentioned DE$\_$DAC, depending on whether the camouflage can be adapted to the scene, we further propose A-DE$\_$DAC. As described in DE$\_$DAC above mentioned, both the textures optimized in the first and second stages are universal. In A-DE$\_$DAC, we try to optimize the global texture on each scene image in the first stage, realizing better adaption to the changing environment. That is, each scene image would be utilized to generate one separate camouflage in the first stage. After that, we try to optimize the local texture on all images. Thus, the global texture optimized by A-DE$\_$DAC in the first stage is sample-independent, while the local texture in the second stage is universal. In this way, the trained local texture is universal to fool detectors, while other areas of objects adopt the global texture that is adaptive to the background environment to improve the camouflage performance of fooling human eyes.

\section{Experiments}\label{sec4}

In this section, extensive experiments are performed to illustrate the effectiveness of our proposed DE$\_$DAC method. In section \ref{dataset}, we describe our experimental setting and details. Section \ref{results} shows the experiment results.

\subsection{Experiment Protocol}\label{dataset}

\subsubsection{Dataset}To approach the physical adversarial camouflage in the real world, the photo-realistic datasets are performed in our experiments. The data preparation is similar to Wang et al. \cite{wang2022fca}. In our experiments, we select the person as the attack object, which is provided in UPC \cite{huang2020universal}. To this end, we select Unreal Engine 4 (UE 4) \cite{qiu2016unrealcv}, a prevalent open-source
simulator for game development research, as the tools to provide various scenes. The UE4 provides a variety of
high-fidelity digital scenarios (e.g., modern urban). The rendered object can be easily combined with the sampled scenes from UE4 by the masks, generating images consisting of objects and background scenes. To compare with previous works, we use UE4 to construct our training and test dataset directly.
The training dataset includes 5000 high-resolution images consisting of the rendered person model and specific scenes,
while the testing set has 500 high-resolution images. The
datasets contain the rendered images that are sampled from different views
angles and distances. In the process of preparing the training dataset, the camera distance of rendering the objects is randomly sampled from [2, 7]. The elevation is randomly sampled from [0, 45], and the azimuth is sampled from [0, 360]. The number of render images is 500, and the number of scene images is set to 10, which is randomly selected from UE 4. Thus we can obtain 5000 images consisting of different render images and scene images. In the test dataset, the camera distance, elevation, and azimuth are randomly selected. The same operation above-mentioned is performed on each specific type of scene. 

\subsubsection{Evaluation Metrics}In this work, we adopt three metrics to evaluate the performance of our proposed camouflage generation method. Two metrics, including attack success rate (ASR) and P@0.5, are utilized to evaluate the adversarial attack performance. ASR is defined as the percentage of the incorrectly detected images on all the correctly detected images. P@0.5 denotes the percentage of the correct detected images on all the images when the detection confident threshd is set to 0.5. Another metric, MSE, is utilized to evaluate the naturalness of the generated camouflage, which is defined as the MSE between the rendered object and the scene images. The lower MSE, the better the generated camouflage towards fooling human eyes.

\subsubsection{Implementation details}

We chose the widely used detector,
Yolo-V3 \cite{redmon2018yolov3}, as our white-box
model to train the adversarial camouflage texture. Then we
evaluate the transferring attack performances
on the following prevalence object detection models:
Yolo-V2 \cite{redmon2017yolo9000}, Yolo-V5 \cite{zhu2021tph}, Faster
R-CNN \cite{ren2015faster}, and Mask R-CNN \cite{he2017mask}. These models are all pretrained on the COCO dataset.
In our experiments, these models are the official
implementation version provided by PyTorch. The detection minimum threshold is set as 0.5 for all detectors by default.
The adversarial camouflage texture is initialized as the random
noise, and the Adam with default parameter is adopted
as the optimizer. The 
learning rate is 0.01. The segmentation network used
to extract the background from the photo-realistic image is
U2-Net \cite{qin2020u2}. In our proposed DE$\_$DAC method, the maximum epochs of the first and second stages are set to 1 and 10, respectively. The number of images in the training datasets in the first and second stages is set to 300 and 5000, respectively. Both batch sizes are set to 1. The $\lambda_{1}$ and $\lambda_{2}$ in $\mathcal{L}_{second}$ loss of the second stage are set to 5$\times e^{-4}$ and 1$\times e^{-7}$ for all the scenes respectively. In the DE algorithm, the parameters setting is as follows. The population size is set to 20. The maximum iteration number $T$ is set to 10. The rate of crossover and mutation is set to 0.6. We conduct all the experiments on a
NVIDIA RTX 3090 24GB GPU cluster.

\subsection{Experimental results}\label{results}

In this section, we conduct a series of experiments to illustrate the effectiveness of our proposed method on the object detection task. The person is selected as our attack object. We conduct the experiments on three types of scenes provided by UE 4, including WinterValley, Forest, and Desert, which are also frequently encountered scenes in real military applications. We study the performance of DAC of the white box attack and black box attack, DE$\_$DAC, and adaptive DE$\_$DAC.  In addition, we also study the performance of two-stage training and different coefficients of DE$\_$DAC. Finaly, we investigate the performance of attacking other objects, such as cars and trucks, to further verify the generality of our proposed method.

\subsubsection{White-box attack of DAC}\label{whiteclass}

In thi section, the performance of white-box attack of DAC is investigated. The Yolo-V3 is selected as the white-box attack model. In our experiments, raw stands for the original images, also called clean images. CAMOU is the camouflage using the method provided by Zhang et al. \cite{2018CAMOU}. AdvCam was first proposed by Dual et al. \cite{duan2020adversarial} to ensure the naturalness and adversarial performance of the generated camouflage using the style transfer technique in the classification task. We transfer the method to the detection task. DAC denotes the optimized texture by our proposed whole method when the number of the local texture is the same as the global texture. The statistical results of p@0.5, ASR, and MSE under different scenes using the Yolo-V3 detection model are listed in Table \ref{person}. The best performance is indicated in \textbf{bold}.

\begin{table*}[htbp]
	\renewcommand\arraystretch{1.5}
	\scriptsize
	\centering
	\caption{Comparison of the P@0.5, ASR and MSE of different attack methods on three person datasets with specific scenes.}
	
	\label{person}
	\setlength{\tabcolsep}{2mm}{
		\begin{tabular}{c|ccccccccc}
			
			\toprule
			
			Methods&	\multicolumn{3}{c}{WinterValley}& \multicolumn{3}{c}{Forest} &  \multicolumn{3}{c}{Desert}  \\
			
			\cmidrule(r){2-4}   \cmidrule(r){5-7}  \cmidrule(r){8-10}
			
			& P@0.5($\downarrow$)	 & ASR($\uparrow$)	 &   MSE($\downarrow$)		
			& P@0.5($\downarrow$)	 & ASR($\uparrow$)	 &   MSE($\downarrow$)
			& P@0.5	($\downarrow$) & ASR($\uparrow$)	 &   MSE($\downarrow$) \\		
			\midrule		
			Raw  &    100.0$\%$   &0.00$\%$ &857.62&100.0$\%$ & 0.00$\%$&729.41 &100.0$\%$&0.00$\%$&1377.10\\	
			{CAMOU }     & 64.8$\%$& 36.1$\%$& 1527.58&78.4$\%$&21.6$\%$& 1393.22& 57.0$\%$&43.0$\%$&2019.04\\
			{AdvCam }     &53.2$\%$& 46.8$\%$& 735.98&24.8$\%$&75.2$\%$&881.13&53.2$\%$&46.8$\%$&1225.15\\

			\midrule
			{DAC(Ours)}     &\textbf{25.4}$\%$ &\textbf{74.6}$\%$ &\textbf{327.00}  &\textbf{23.8}$\%$ &\textbf{76.2}$\%$&\textbf{298.61}&\textbf{10.0}$\%$&\textbf{90.0}$\%$&\textbf{524.63}\\

			\bottomrule		
	\end{tabular}}
\end{table*}

\begin{figure*}[htbp]
	\centering
	\subfigure[Clean]{
		\includegraphics[height=0.2\linewidth]{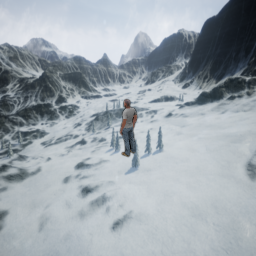}
	}
	\subfigure[CAMOU]{
		\includegraphics[height=0.2\linewidth]{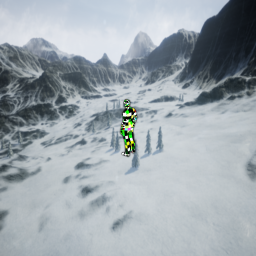}
	}
	\subfigure[AdvCam]{
		\includegraphics[height=0.2\linewidth]{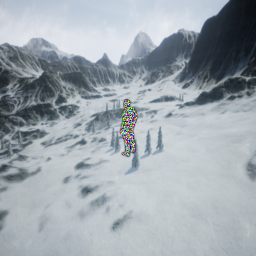}
	}
	\subfigure[DAC]{
		\includegraphics[height=0.2\linewidth]{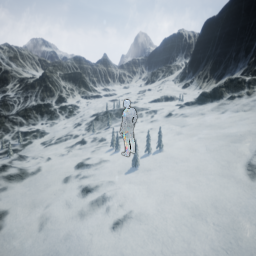}
	}

	\subfigure[Clean]{
		\includegraphics[height=0.2\linewidth]{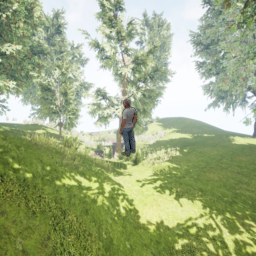}
	}
	\subfigure[CAMOU]{
		\includegraphics[height=0.2\linewidth]{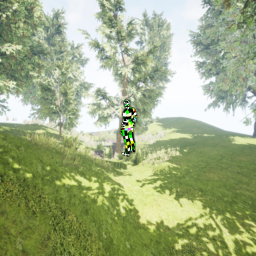}
	}
	\subfigure[AdvCam]{
		\includegraphics[height=0.2\linewidth]{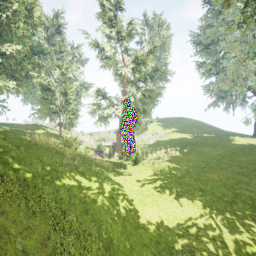}
	}
	\subfigure[DAC]{
		\includegraphics[height=0.2\linewidth]{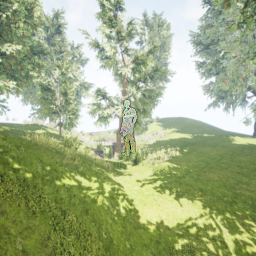}
	}
	
	\subfigure[Clean]{
		\includegraphics[height=0.2\linewidth]{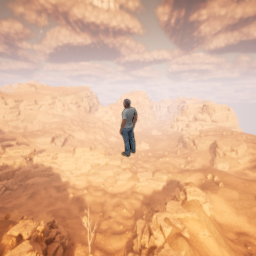}
	}
	\subfigure[CAMOU]{
		\includegraphics[height=0.2\linewidth]{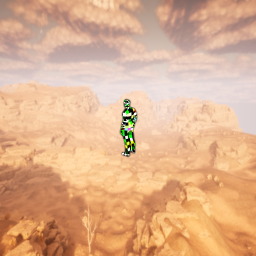}
	}
	\subfigure[AdvCam]{
		\includegraphics[height=0.2\linewidth]{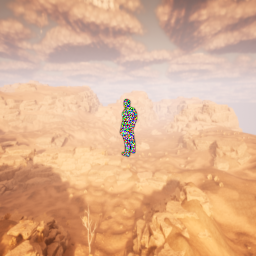}
	}
	\subfigure[DAC]{
		\includegraphics[height=0.2\linewidth]{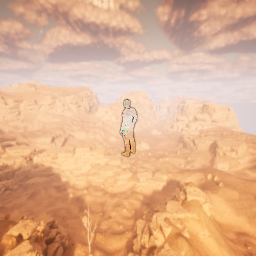}
	}
	
	\caption{The visualization of different adversarial attack methods on WinterValley, Forest, and Desert scenes.}
	\label{nsga}
\end{figure*}

From Table \ref{person}, we can see that camouflage optimized by DAC can obtain almost the lowest MSE among all methods on three datasets and possess certain adversarial performance simultaneously, causing the significant drop in p@0.5 and increase in ASR than clean images. In addition, on three datasets, all the p@0.5 and ASR values of DAC are the best among the comparison methods. In contrast, the MSE value of generated camouflage by CAMOU is relatively too large, causing the conspicuous appearance towards human eyes, which also can be seen in other methods. From the results of AdvCam, we can see that compared with the classification task, it is more difficult to directly train the combination of three loss functions using the universal dataset for the detection task. The trained camouflage texture can not obtain a good trade-off between fooling human eyes and object detectors. The camouflages generated by us and existing methods towards different scenes are presented in Figure~\ref{nsga}. From the aspect of fooling human eyes, pure camouflage and DE$\_$DAC can obtain better performance than other methods.

\subsubsection{Black-box attack of DAC}\label{boxclass}

To further illustrate the effectiveness of our proposed method, we conduct the transfer experiment on other object detectors, including Yolo-V2, Yolo-V5, FasterRCNN, and MaskRCNN. The performance of the generated camouflage by Yolo-V3 is evaluated on the above-mentioned models. Table \ref{otherclass1} shows the ASR and P@0.5 values of detection results by different detectors on three datasets, respectively. 

\begin{table*}[htbp]
	\renewcommand\arraystretch{1.5}
	\scriptsize
	\centering
	\caption{Comparison of the P@0.5, ASR and MSE of different attack methods on other object detectors in three specific scenes.}
	
	\label{otherclass1}
	\setlength{\tabcolsep}{2mm}{
		\begin{tabular}{c|c|cccccccc}
			
			\toprule
			
			Scene&	Methods&	\multicolumn{2}{c}{Yolo-V2}& \multicolumn{2}{c}{Yolo-V5} &  \multicolumn{2}{c}{FasterRCNN} &  \multicolumn{2}{c}{MaskRCNN}  \\
			
			\cmidrule(r){3-4}   \cmidrule(r){5-6}  \cmidrule(r){7-8} \cmidrule(r){9-10}
			
			&	& Accuracy($\downarrow$) & ASR ($\uparrow$)	 &   Accuracy($\downarrow$)		
			& ASR ($\uparrow$)	 & Accuracy($\downarrow$)	 & ASR($\uparrow$)
			& Accuracy($\downarrow$)	 & ASR($\uparrow$)	   \\		
			\midrule		
			WinterValley  &   Raw  &     84.4$\%$&0.00$\%$ &100$\%$&0.00$\%$&100$\%$&0.00$\%$&100$\%$&0.00$\%$\\	
			&{CAMOU }     & 44.2$\%$ &47.6$\%$ &28.4$\%$ &71.6$\%$&84.8$\%$&15.2$\%$&94.8$\%$&5.2$\%$\\
			&	{AdvCam }     &19.4$\%$&77.0$\%$&67.8$\%$&32.2$\%$&69.0$\%$&31.0$\%$&77.2$\%$&22.8$\%$\\
			& 	{DAC(Ours)}     & \textbf{17.8}$\%$&\textbf{78.9}$\%$&\textbf{6.8}$\%$&\textbf{93.2}$\%$& \textbf{58.2}$\%$&\textbf{41.8}$\%$&\textbf{62.6}$\%$&\textbf{37.4}$\%$\\
			\midrule		
			Forest  &  Raw  &  86.6$\%$ &0.00$\%$&100$\%$& 0.00$\%$ &100$\%$ &0.00$\%$ &100$\%$&0.00$\%$\\	
			&	{CAMOU }    & 53.8$\%$  &37.9$\%$&16.8$\%$&83.2$\%$&71.4$\%$&28.6$\%$&80.6$\%$&19.4$\%$\\
			&	{AdvCam }     &36.6$\%$&57.7$\%$&77.0$\%$&23.0$\%$&57.4$\%$&42.6$\%$&64.4$\%$&35.6$\%$\\
			& 	{DAC(Ours)}     &\textbf{30.2}$\%$&\textbf{65.1}$\%$&\textbf{11.4}$\%$&\textbf{88.6}$\%$&\textbf{51.4}$\%$&\textbf{48.6}$\%$&\textbf{46.0}$\%$&\textbf{54.0}$\%$\\	
			
			\midrule		
			Desert  &  Raw&   83.8$\%$  &0.00$\%$&100$\%$&0.00$\%$&100$\%$&0.00$\%$ &100$\%$& 0.00$\%$\\	
			&	{CAMOU }     &62.6$\%$ &33.3$\%$&17.6$\%$&30.6$\%$& 90.6$\%$ &9.40$\%$&89.4$\%$ & 10.6$\%$\\
			&	{AdvCam }     &29.6$\%$& 23.1$\%$&44.8$\%$&55.2$\%$ &67.0$\%$&33.0$\%$&64.6$\%$&35.4$\%$\\
			& 	{DAC(Ours)}     &\textbf{36.0}$\%$ &\textbf{57.0}$\%$& \textbf{12.8}$\%$&\textbf{87.2}$\%$  &\textbf{77.4}$\%$&\textbf{22.6}$\%$&\textbf{52.2}$\%$&\textbf{47.8}$\%$\\		
			\bottomrule		
	\end{tabular}}
\end{table*}

\begin{figure*}[htbp]
	\centering
	\subfigure[Clean]{
		\includegraphics[height=0.145\linewidth]{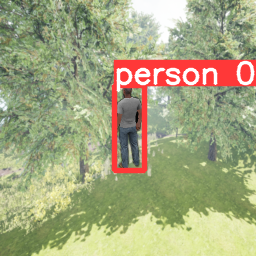}
	}
	\subfigure[Clean]{
		\includegraphics[height=0.145\linewidth]{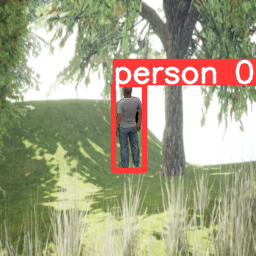}
	}
	\subfigure[Clean]{
		\includegraphics[height=0.145\linewidth]{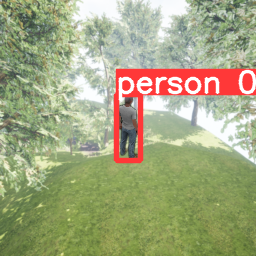}
	}
	\subfigure[Clean]{
		\includegraphics[height=0.145\linewidth]{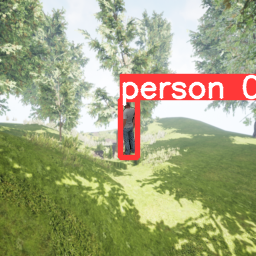}
	}
	\subfigure[Clean]{
		\includegraphics[height=0.145\linewidth]{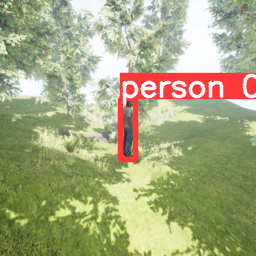}
	}
	\subfigure[Clean]{
		\includegraphics[height=0.145\linewidth]{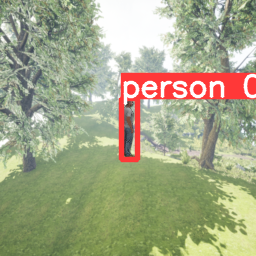}
	}
	
	\subfigure[CAMOU]{
	\includegraphics[height=0.145\linewidth]{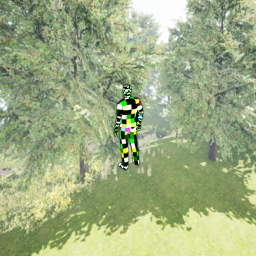}
}
\subfigure[CAMOU]{
	\includegraphics[height=0.145\linewidth]{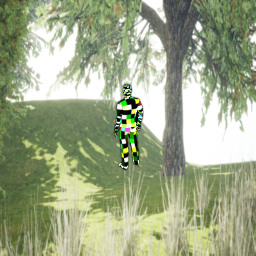}
}
\subfigure[CAMOU]{
	\includegraphics[height=0.145\linewidth]{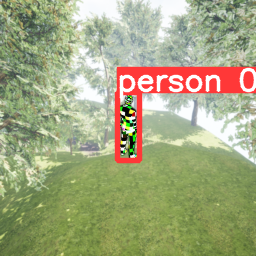}
}
\subfigure[CAMOU]{
	\includegraphics[height=0.145\linewidth]{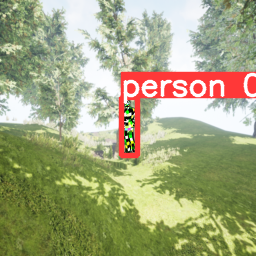}
}
\subfigure[CAMOU]{
	\includegraphics[height=0.145\linewidth]{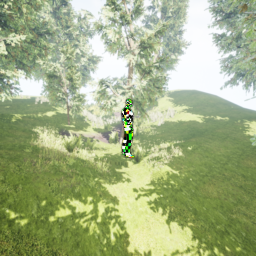}
}
\subfigure[CAMOU]{
	\includegraphics[height=0.145\linewidth]{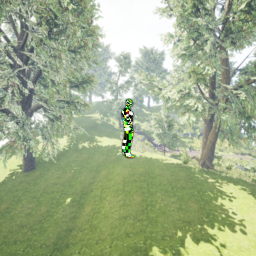}
}
	
	\subfigure[AdvCam]{
	\includegraphics[height=0.145\linewidth]{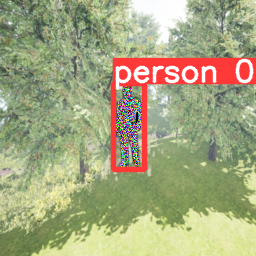}
}
\subfigure[AdvCam]{
	\includegraphics[height=0.145\linewidth]{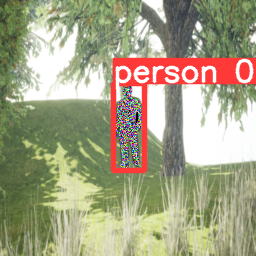}
}
\subfigure[AdvCam]{
	\includegraphics[height=0.145\linewidth]{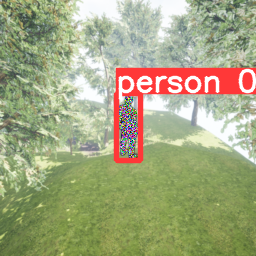}
}
\subfigure[AdvCam]{
	\includegraphics[height=0.145\linewidth]{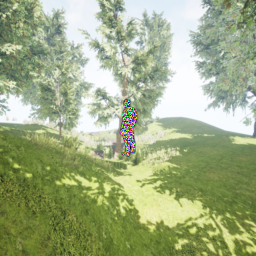}
}
\subfigure[AdvCam]{
	\includegraphics[height=0.145\linewidth]{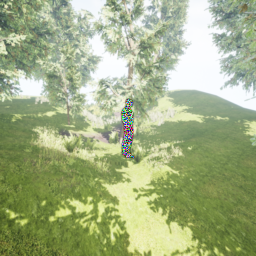}
}
\subfigure[AdvCam]{
	\includegraphics[height=0.145\linewidth]{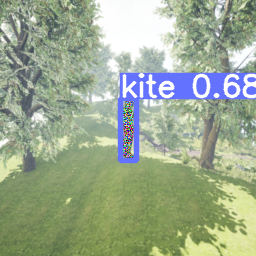}
}
%

\subfigure[DAC]{
	\includegraphics[height=0.145\linewidth]{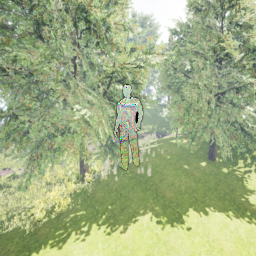}
}
\subfigure[DAC]{
	\includegraphics[height=0.145\linewidth]{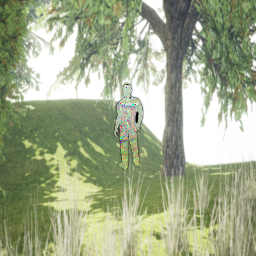}
}
\subfigure[DAC]{
	\includegraphics[height=0.145\linewidth]{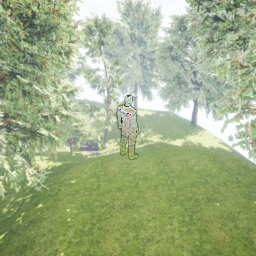}
}
\subfigure[DAC]{
	\includegraphics[height=0.145\linewidth]{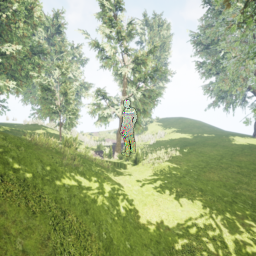}
}
\subfigure[DAC]{
	\includegraphics[height=0.145\linewidth]{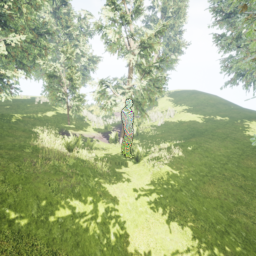}
}
\subfigure[DAC]{
	\includegraphics[height=0.145\linewidth]{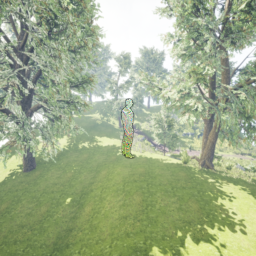}
}
	
	\caption{The visualization of different adversarial attack methods on Forest scene.}
	\label{comparison}
\end{figure*}

From Table \ref{otherclass1}, we can see that among four detectors, the p@0.5 value of Yolo-V2 is about 85$\%$, and that of other detectors is 100$\%$, which shows superior detection performance. However, these superior detectors can still be fooled by our proposed method. On all object detectors, compared with existing methods, DAC obtains the highest ASR and lowest p@0.5 values. For example, the ASR of attacking Yolo-V5 on the WinterValley scene using DAC can reach 93.2$\%$, while that of CAMOU and AdvCam are just 71.6$\%$ and 32.2$\%$, respectively. In most detection cases, the ASR of DAC can also be over 50.0$\%$. Some examples generated by different methods are visualized in Figure \ref{comparison}, which are detected by the Yolo-V5 model. Under the same training and test settings, we can see that our proposed method can obtain the lowest detection performance from different views. Thus our proposed DAC method can also possess competitive transfer adversarial performance under black-box attack settings.

%

\begin{figure*}[htbp]
	\centering
	
	\subfigure[WinterValley]{
		\includegraphics[height=0.37\linewidth]{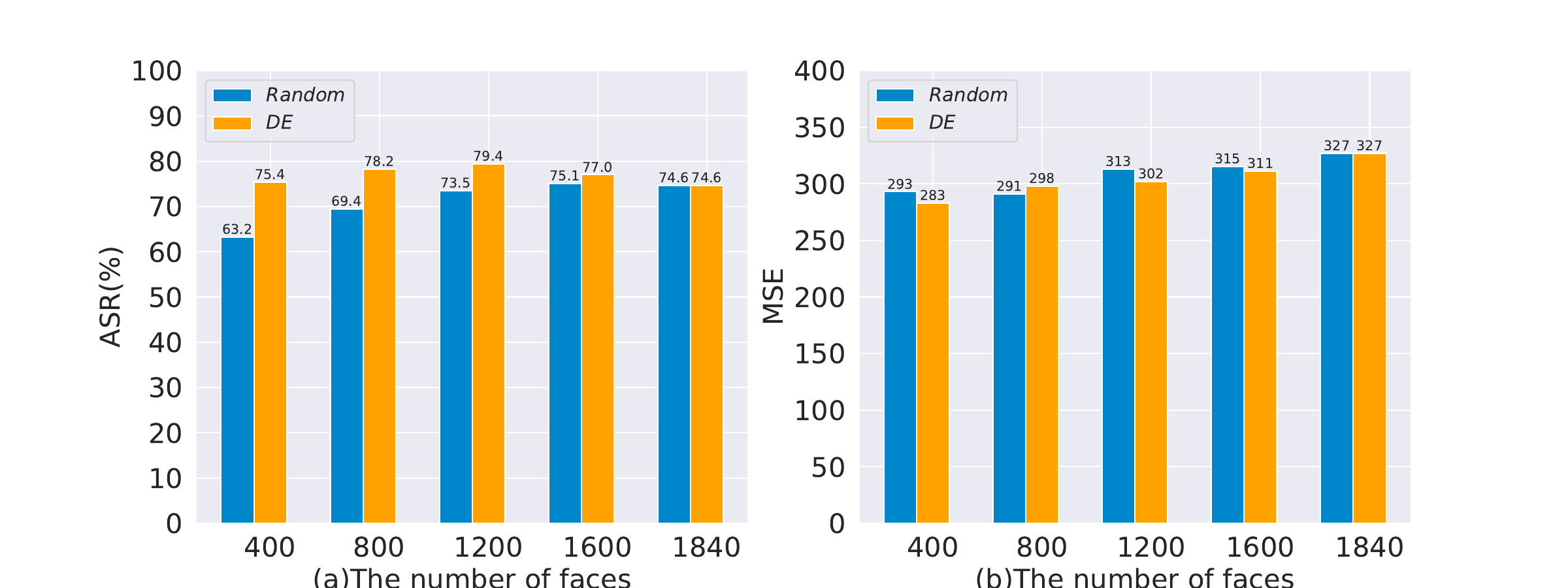}
	}
	
	\subfigure[Forest]{
		\includegraphics[height=0.37\linewidth]{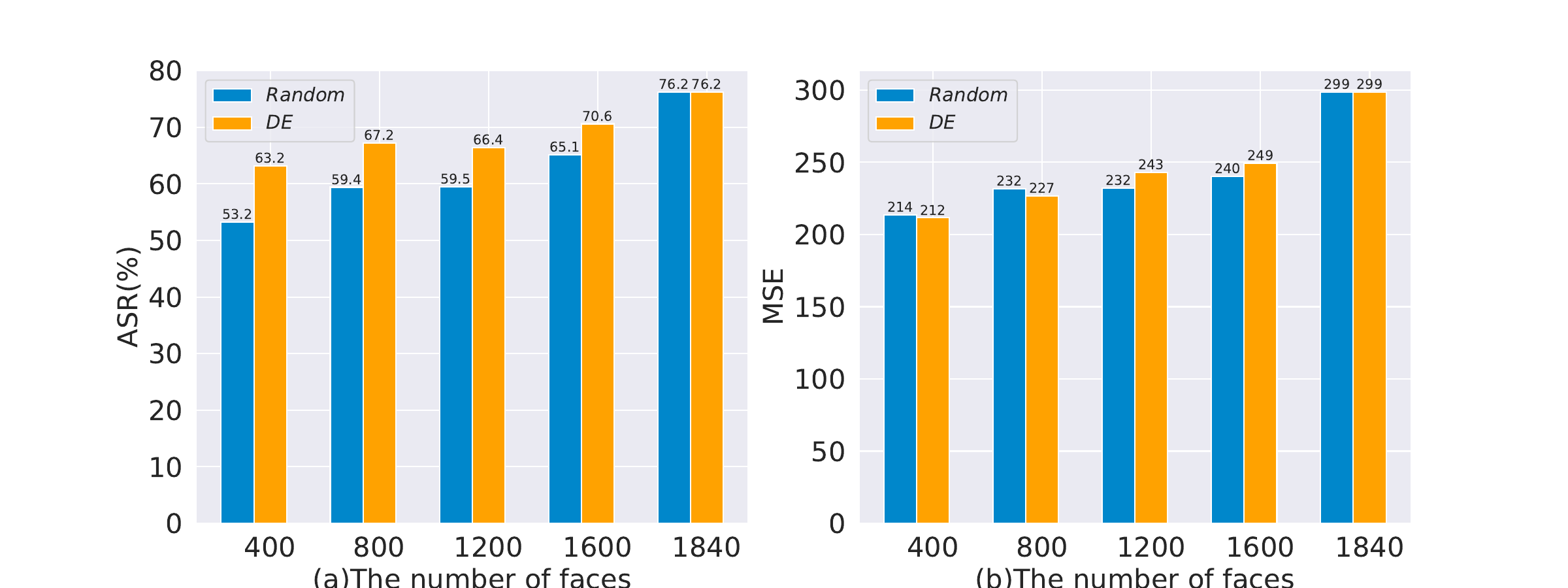}
	}
	
	\subfigure[Desert]{
		\includegraphics[height=0.37\linewidth]{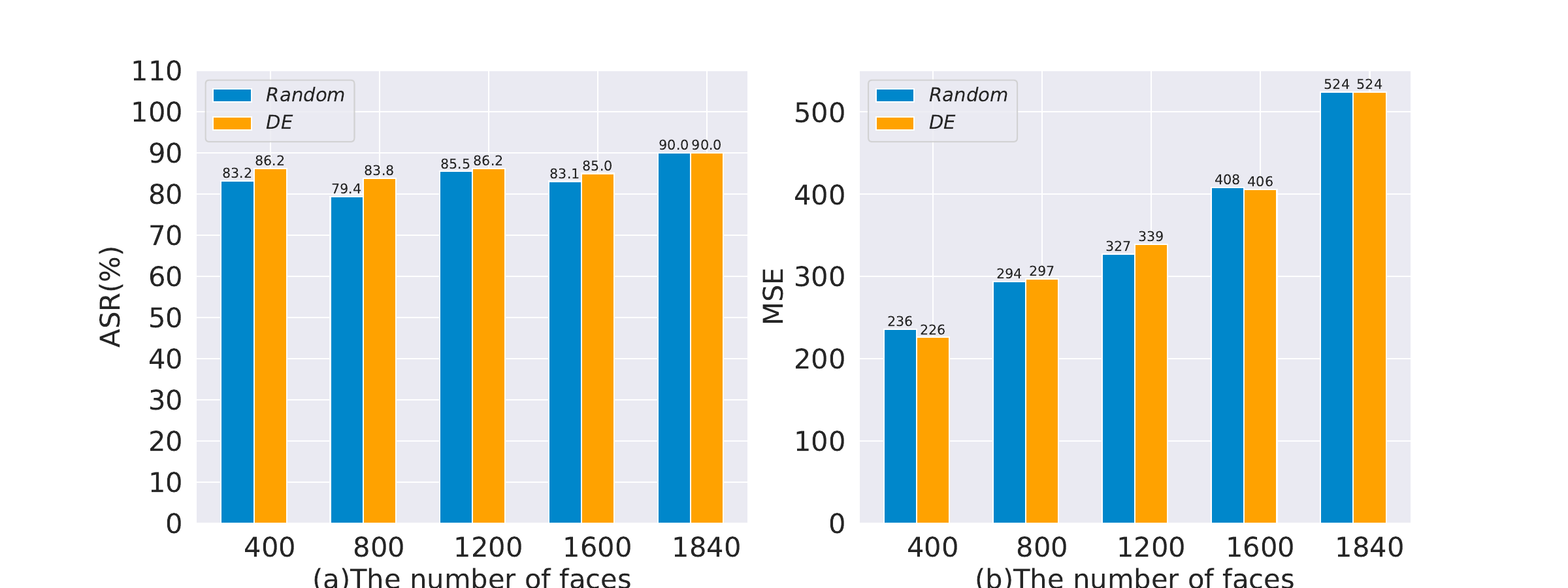}
	}
	\caption{The performance of ASR and MSE of random search and DE with different number of face in different scenes.}
	\label{de}
\end{figure*}

\subsubsection{The performance of DE$\_$DAC}\label{de_dac}

In this section, under the case that the number of optimized local texture is limited, we investigate the performance improvement brought by the DE algorithm introduced in the proposed DE$\_$DAC. In our experiment, the number of faces of the person model is 1840. Thus the full surface can be represented by 1840 inter values from 1 to 1840. We can select unduplicated $n_{f}$ values from 1 to 1840 to obtain the corresponding area in the person model. If $n_{f}$ does not reach 1840, our proposed DE$\_$DAC can search for the near-optimal attack area under the settled number of faces. To illustrate the effectiveness of our proposed method, we set the number of faces as 400, 800, 1200, 1600, and 1840, respectively, and utilize the random selection strategy and DE algorithm to select the faces of the optimized local texture. The comparisons of the p@0.5 and ASR of the generated camouflage by random selection and DE are presented in Figure \ref{de}, respectively.

From Figure \ref{de}, we can see that with the increasing number of faces, the attack performance, including p@0.5 and ASR, also increases. The smaller number of faces means less perturbation and a more natural appearance. Given the settled number of faces, DE can find a better attack area compared with random selection, which has a higher ASR and a lower p@0.5. From Figure \ref{de} (b), we can see that the ASR of camouflage by random selection is only 59.4$\%$ when the settled number of faces is 800, while the ASR of camouflage by DE can reach 67.2$\%$. Particularly, the ASR of camouflage by random search with the number of faces 1600 is just 65.1$\%$. It means that the adversarial performance of camouflages using 800 faces searched by DE can outperform that using 1600 faces by random search. On the one hand, it shows that our proposed DE can effectively find the important area to improve adversarial performance. On the other hand, it shows the effectiveness of our proposed DE$\_$DAC method, possessing competitive adversarial performance even though a small number of faces are given.

\begin{figure*}[htbp]
	\centering
	\subfigure[400]{
		\includegraphics[height=0.18\linewidth]{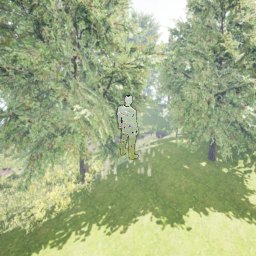}
	}
	\subfigure[800]{
		\includegraphics[height=0.18\linewidth]{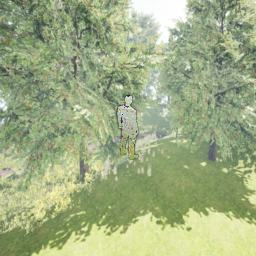}
	}
	\subfigure[1200]{
		\includegraphics[height=0.18\linewidth]{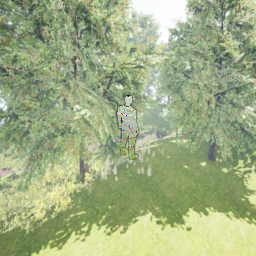}
	}
	\subfigure[1600]{
		\includegraphics[height=0.18\linewidth]{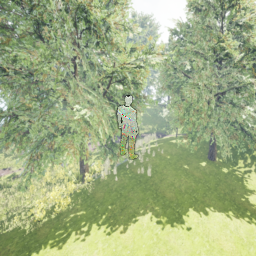}
	}
	\subfigure[1840]{
		\includegraphics[height=0.18\linewidth]{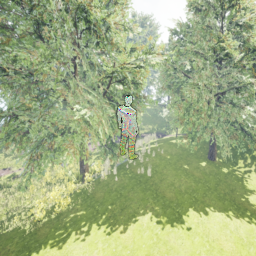}
	}	
	\caption{The visualization of the searched attack area with different number of faces in the second stage.}
	\label{pattern}
\end{figure*}

\begin{figure*}[htbp]
	\centering
	
	\subfigure[WinterValley]{
	\includegraphics[height=0.37\linewidth]{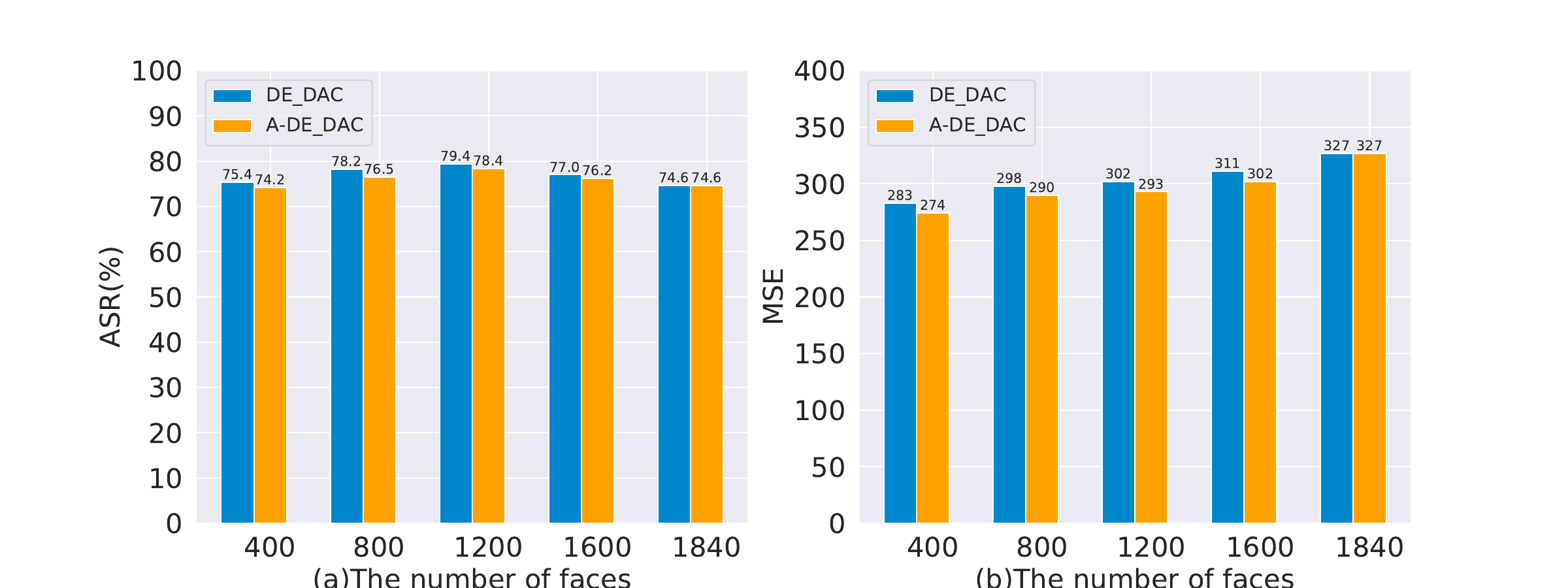}
}

\subfigure[Forest]{
	\includegraphics[height=0.37\linewidth]{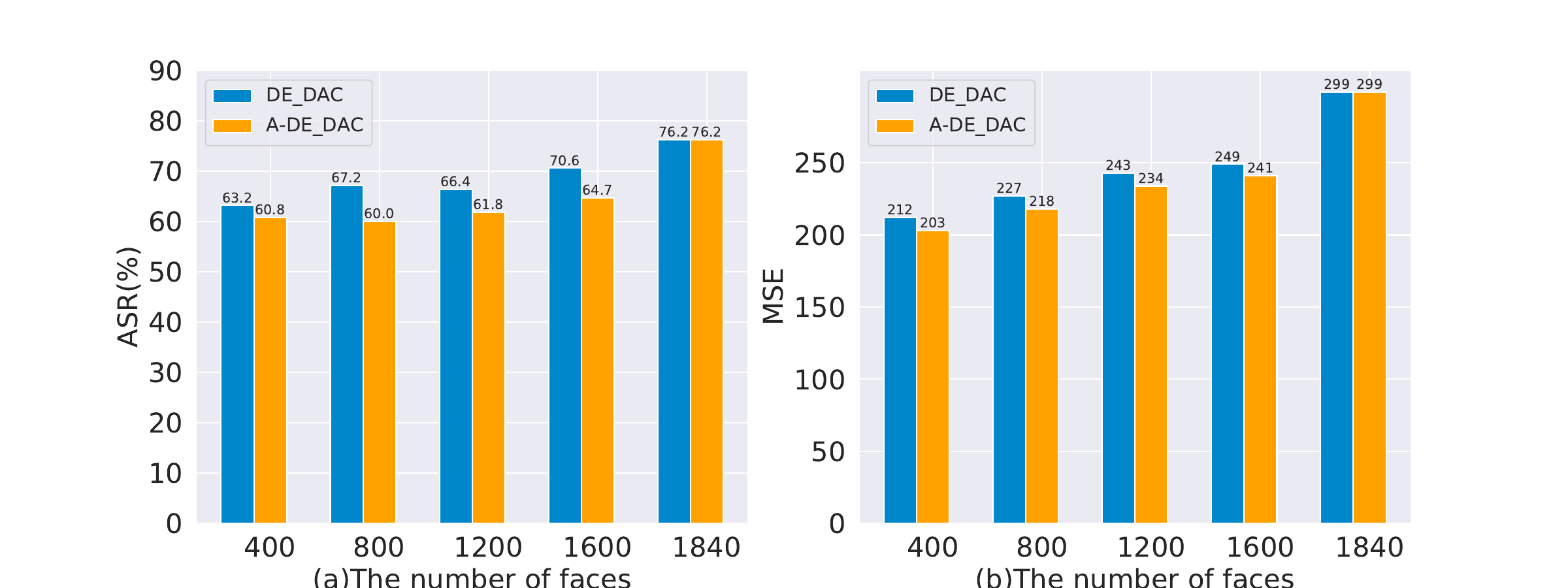}
}

\subfigure[Desert]{
	\includegraphics[height=0.37\linewidth]{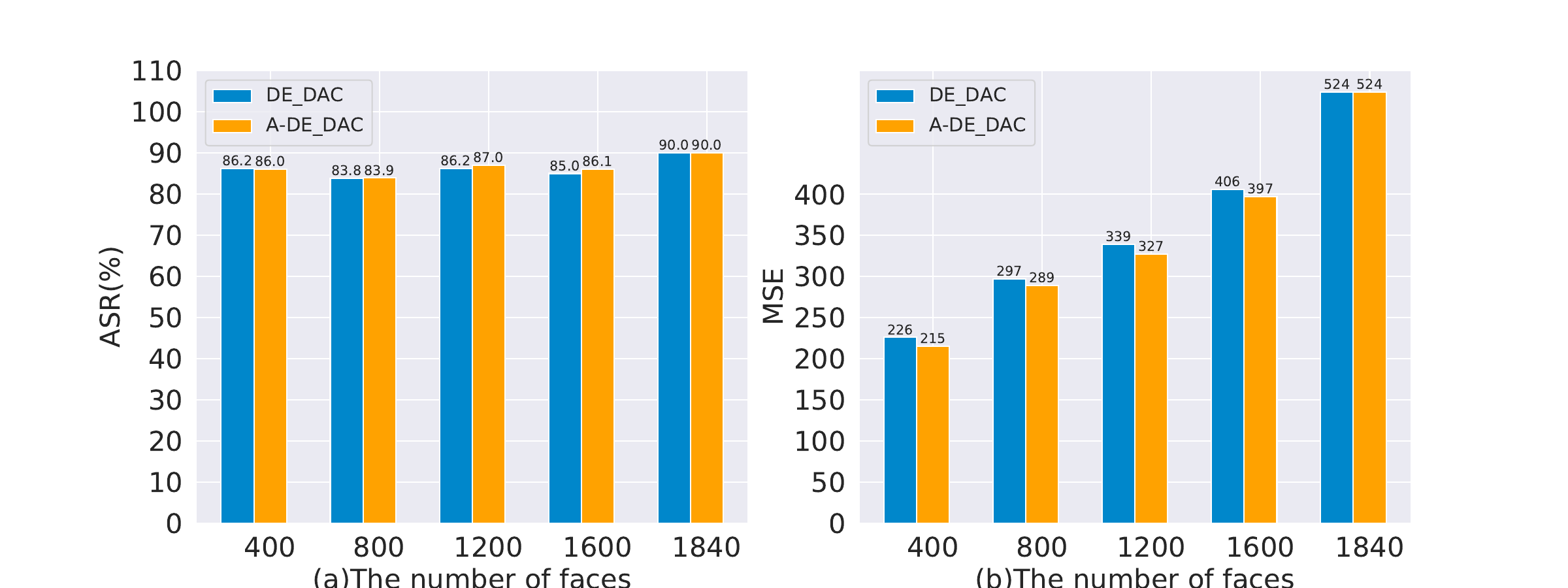}
}
	\caption{The performance of ASR and MSE of DE$\_$DAC and A-DE$\_$DAC with different number of faces in different scenes.}
	\label{Ade}
\end{figure*}

\subsubsection{The performance of Adaptive DE$\_$DAC}\label{ADde_dac}
In this section, we study the performance of the proposed adaptive DE$\_$DAC method. Specifically, taking the forest scene as an example, we first train the texture according to each scene image belonging to the forest scene in the first stage of DE$\_$DAC, obtaining ten trained global textures. In the second stage, we select the local texture to optimize. As for each input image, the global texture is selected from the corresponding one. Thus in the whole training process, the optimized local texture is universal while the global texture is adaptive. The procedure is the same in the evaluation of the final textures. We also set the different numbers of faces in the selection of local textures, including 400, 800, 1200, 1600, and 1840. The performance of A-DE$\_$DAC and DE$\_$DAC under three datasets is presented in Figure \ref{Ade}.

\begin{figure*}[htbp]
	\centering
	\subfigure[]{
		\includegraphics[height=0.18\linewidth]{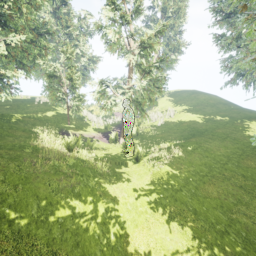}
	}
	\subfigure[]{
		\includegraphics[height=0.18\linewidth]{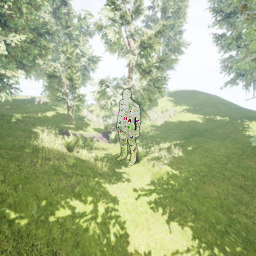}
	}
	\subfigure[]{
		\includegraphics[height=0.18\linewidth]{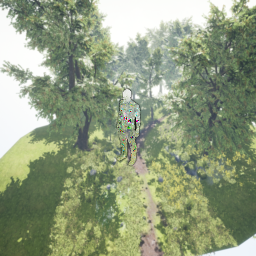}
	}
	\subfigure[]{
		\includegraphics[height=0.18\linewidth]{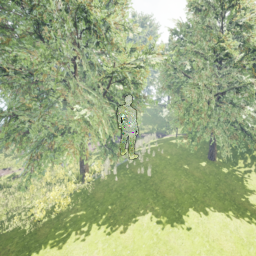}
	}
	\subfigure[]{
		\includegraphics[height=0.18\linewidth]{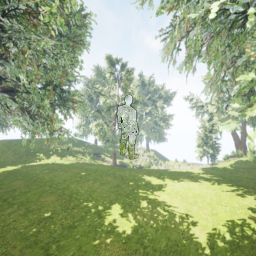}
	}	
	\caption{The visualization of adaptive DE$\_$DAC.}
	\label{adaptive}
\end{figure*}

From Figure \ref{Ade}, we can see that compared with DE$\_$DAC, A-DE$\_$DAC can obtain better naturalness with the scenes. On three datasets, all of the MSE of optimized textures are lower than  DE$\_$DAC when the number of faces is the same, dropping about 10. It can be predictable that A-DE$\_$DAC can fastly generate the global texture adaptively according to the background of each image. In general, the MSE value with the increasing number of faces becomes larger in A-DE$\_$DAC, which also corresponds to the previous conclusions. In contrast, in most cases, the performance of ASR of A-DE$\_$DAC is slightly worse than DE$\_$DAC. It may be because the optimization problem of  A-DE$\_$DAC is more difficult than DE$\_$DAC under our settings. On the whole, A-DE$\_$DAC can obtain better camouflage towards human eyes with a slight sacrifice of ASR. Figure \ref{adaptive} shows the visualization of generated AEs by A-DE$\_$DAC. The generated global camouflage is adaptive to the background, while the trained local camouflage is universal.

\subsubsection{The performance of two-stage training}
To verify the effectiveness of our proposed two-stage training method, we conduct the ablation study. We first make the comparison with the one-stage method, denoted as one-stage DAC. One stage means that the camouflage is trained jointly by the adversarial loss, smooth loss and the MSE between the rendered with the scenes, which is also corresponding to the second stage of the proposed method. Typical one-stage method has been adopted in the classification task such as Dual et al. \cite{duan2020adversarial}. In addition, we also compare our method with the pure camouflage, which is the optimized texture after the first stage in our proposed method. We evaluate the white-box attack performance of one-stage DAC and pure camouflage under three scenes. The Yolo-V3 is selected as the threat model. The statistical resulsts about p@0.5, ASR, and MSE are presented in Table \ref{study}.

\begin{figure*}[htbp]
 	\centering
 	\subfigure[Clean]{
 		\includegraphics[height=0.2\linewidth]{clean_40.png}
 	}
 	\subfigure[Pure Camouflage]{
 		\includegraphics[height=0.2\linewidth]{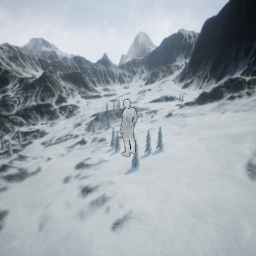}
 	}
 	\subfigure[One-stage DAC]{
 		\includegraphics[height=0.2\linewidth]{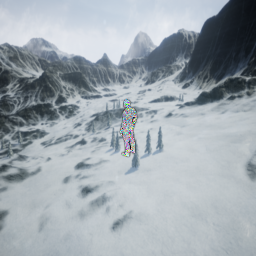}
 	}
 	\subfigure[DAC]{
 		\includegraphics[height=0.2\linewidth]{DAC_40.png}
 	}
 	
 	\subfigure[Clean]{
 		\includegraphics[height=0.2\linewidth]{clean_40_forest.png}
 	}
 	\subfigure[Pure Camouflage]{
 		\includegraphics[height=0.2\linewidth]{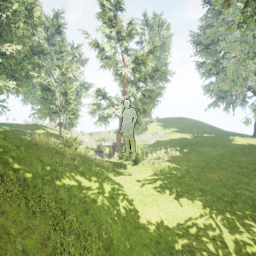}
 	}
 	\subfigure[One-stage DAC]{
 		\includegraphics[height=0.2\linewidth]{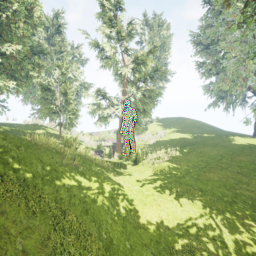}
 	}
 	\subfigure[DAC]{
 		\includegraphics[height=0.2\linewidth]{DAC_40_forest.png}
 	}
 	
 	\subfigure[Clean]{
 		\includegraphics[height=0.2\linewidth]{clean_40_desert.png}
 	}
 	\subfigure[Pure Camouflage]{
 		\includegraphics[height=0.2\linewidth]{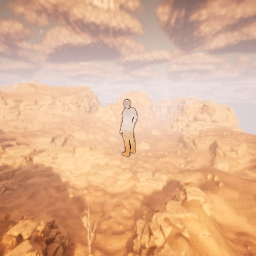}
 	}
 	\subfigure[One-stage DAC]{
 		\includegraphics[height=0.2\linewidth]{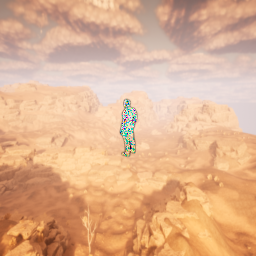}
 	}
 	\subfigure[DAC]{
 		\includegraphics[height=0.2\linewidth]{DAC_40_desert.png}
 	}
 	
 	\caption{The visualization of pure camouflage, one-stage DAC, and DAC on WinterValley, Forest, and Desert scenes.}
 	\label{abstudy}
 \end{figure*}

\begin{table*}[htbp]
	\renewcommand\arraystretch{1.5}
	\scriptsize
	\centering
	\caption{Comparison of the P@0.5, ASR and MSE of different attack methods on three person datasets with specific scenes.}
	
	\label{study}
	\setlength{\tabcolsep}{2mm}{
		\begin{tabular}{c|ccccccccc}
			
			\toprule
			
			Methods&	\multicolumn{3}{c}{WinterValley}& \multicolumn{3}{c}{Forest} &  \multicolumn{3}{c}{Desert}  \\
			
			\cmidrule(r){2-4}   \cmidrule(r){5-7}  \cmidrule(r){8-10}
			
			& P@0.5($\downarrow$)	 & ASR($\uparrow$)	 &   MSE($\downarrow$)		
			& P@0.5($\downarrow$)	 & ASR($\uparrow$)	 &   MSE($\downarrow$)
			& P@0.5	($\downarrow$) & ASR($\uparrow$)	 &   MSE($\downarrow$) \\		
			\midrule
			Raw  &    100.0$\%$   &0.00$\%$ &857.62&100.0$\%$ & 0.00$\%$&729.41 &100.0$\%$&0.00$\%$&1377.10\\		
			{Pure Camouflage}     &72.6$\%$ &27.4$\%$ &\textbf{252.22} &68.8$\%$&31.2$\%$&\textbf{168.98}&29.8$\%$&70.2$\%$&\textbf{118.23}\\
			{One-stage DAC}  & 30.6$\%$   &69.4$\%$&407.85&30.6$\%$ &69.4$\%$ &388.74&15.6$\%$&84.4$\%$&832.86\\
			\midrule
			
			{DAC(Ours)}     &\textbf{25.4}$\%$ &\textbf{74.6}$\%$ &327.00  &\textbf{23.8}$\%$ &\textbf{76.2}$\%$&298.61&\textbf{10.0}$\%$&\textbf{90.0}$\%$&524.63\\						
			\bottomrule		
	\end{tabular}}
\end{table*}
 From Table \ref{study}, we can see that the MSE of DAC is close to the pure camouflage optimized by the first stage. It indicates that even though introducing the local texture in the second stage to optimze to fool DNN models, we do not impair the naturalness towards human eyes in our proposed two-stage training. In addition, we can see that pure camouflage directly optimizes the global texture to fool human eyes, achieving a certain adversarial performance and making the accuracy of detectors decrease from 100$\%$ to about 70$\%$. It shows that the optimized camouflage by the first stage can provide a better initialization for the second stage, further improving the final adversarial performance of DAC. That is, our proposed two stage dual adversarial camouflage can collaboratively boost performances, where the first stage makes up for the disadvantage of not being natural in the second stage, and the second stage bridges the gap of fooling object detectors in the first stage. The effectiveness of the proposed two-stage training can also be verified in the comparison between DAC and one-stage DAC. It can be seen that DAC performs better than one-stage DAC on all metrics, including p@0.5, ASR, and MSE on all cases. The visualization of the camouflage generated by different methods is presented in Figure \ref{abstudy}. These experimental results show that our proposed method strikes a better balance between fooling the detector and human eyes than other methods.
 
\subsubsection{The performance of the coefficient $\lambda_{1}$}\label{coefficients}

\begin{figure*}[htbp]
	\centering
	\subfigure[ASR]{
		\includegraphics[height=0.3\linewidth]{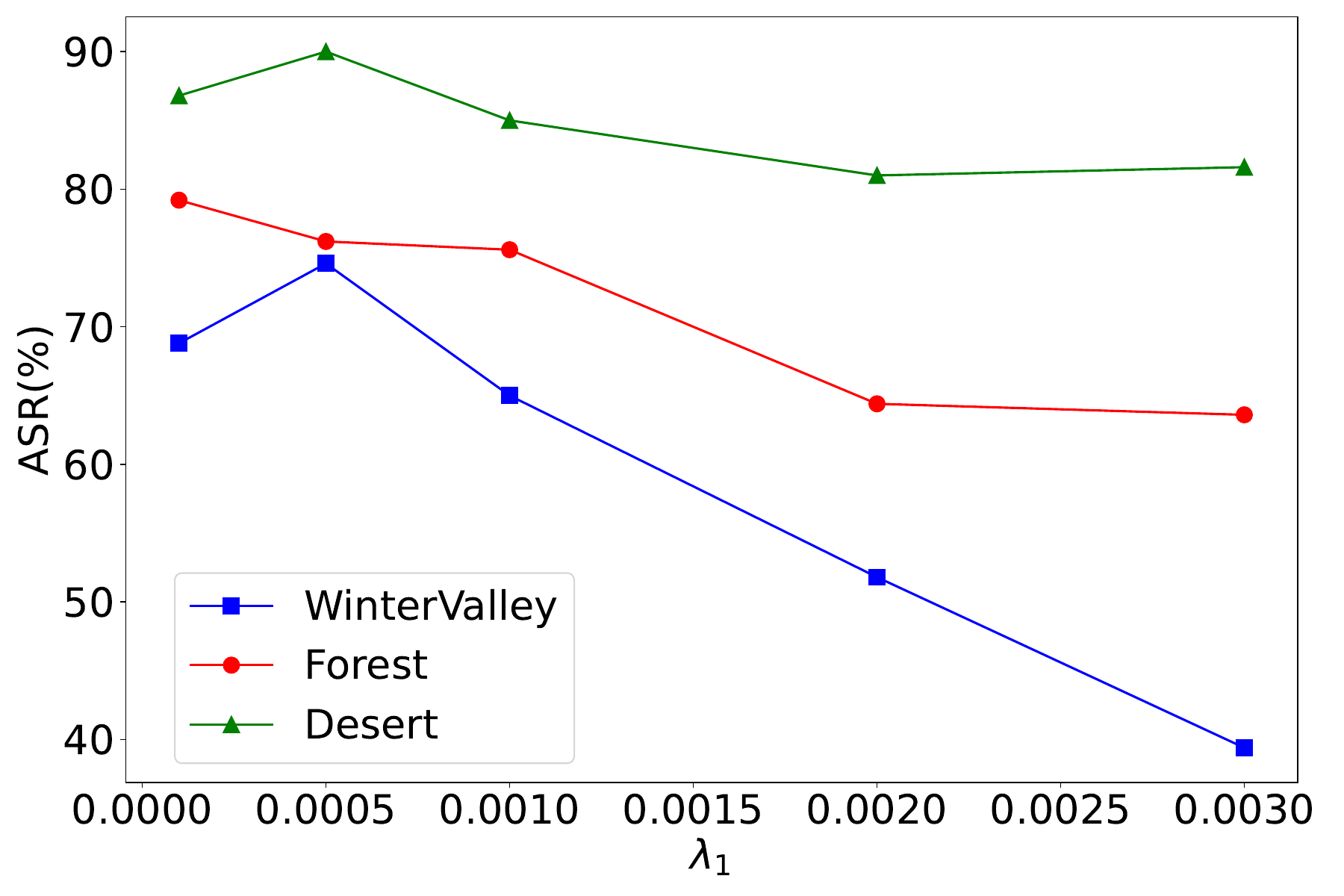}
	}
	\subfigure[MSE]{
		\includegraphics[height=0.3\linewidth]{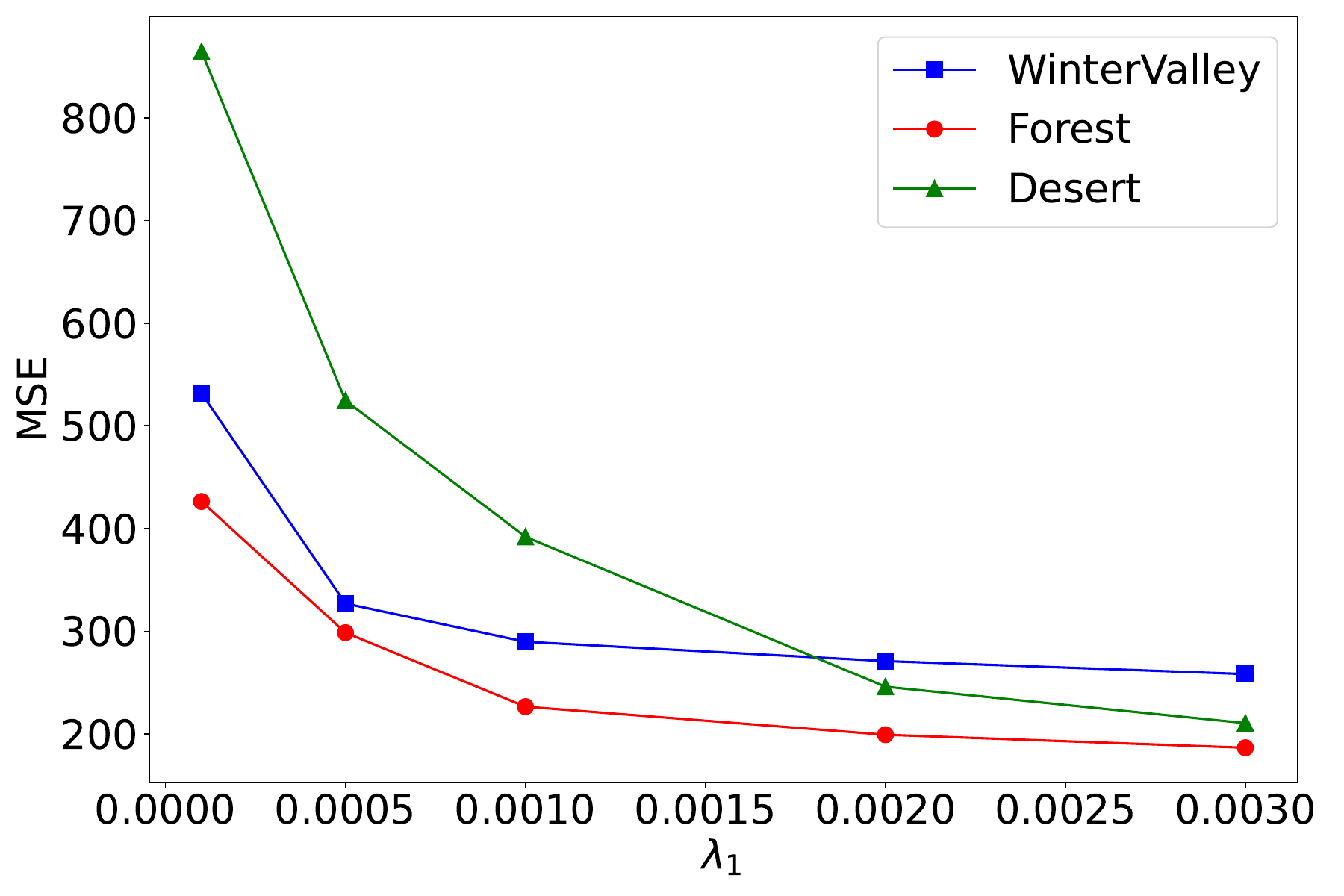}
	}	
	\caption{The effect of $\lambda_{1}$ on ASR and MSE on three datasets.}
	\label{lambda_1}
\end{figure*}

The coefficient $\lambda_{1}$ in the total loss plays a key role in the final performance of our optimized camouflage. To study the effect of $\lambda_{1}$ on the performance of ASR and MSE of our proposed method, we conduct the experiment on three datasets under different $\lambda_{1}$  settings. In our experiments,  $\lambda_{1}$ is set to 0.0001, 0.0005, 0.01, 0.02 and 0.03 respectively, while $\lambda_{2}$ remains unchanged. The selected number of faces in the second stage is set to 1840. We can train five different camouflages under five parameter settings by attacking the Yolo-V3 detector. All the generated test AEs are also detected by Yolo-V3. The statistical result of ASR and MSE of five camouflages is presented in Figure \ref{lambda_1}. From Figure \ref{lambda_1}, we can see that the selection of $\lambda_{1}$ has a great influence on the performance of ASR and MSE. With the increase of $\lambda_{1}$, both the ASR and MSE decrease, which means the adversarial performance and the naturalness of the generated camouflage are contradictory. In addition, as the $\lambda_{1}$ increases, the change in ASR becomes progressively greater, while the change in MSE tends to flatten out. Thus, we can select a relatively better value like 0.0005 to keep a good balance between the adversarial performance and naturalness towards human eyes. In general, the ASR using different $\lambda_{1}$ can be maintained at a high level. For instance, even though the $\lambda_{1}$ is set to 0.003, the ASR on Forest and Desert datasets can still be over 65$\%$ and 82$\%$, which verifies the effectiveness of our proposed two-stage DAC method.

\subsubsection{Other objects and scenes}\label{otherobjects}

\begin{figure}[h]
	\centering
	\subfigure[Camouflage]{
		\includegraphics[height=0.145\linewidth]{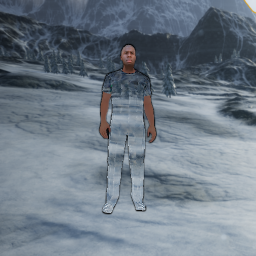}
	}
	\subfigure[Camouflage]{
		\includegraphics[height=0.145\linewidth]{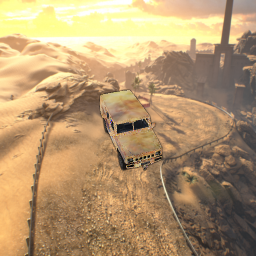}
	}
	\subfigure[Camouflage]{
		\includegraphics[height=0.145\linewidth]{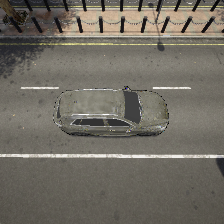}
	}	
	\subfigure[Camouflage]{
		\includegraphics[height=0.145\linewidth]{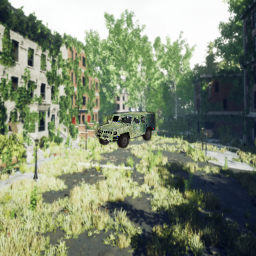}
	}
	\subfigure[Camouflage]{
		\includegraphics[height=0.145\linewidth]{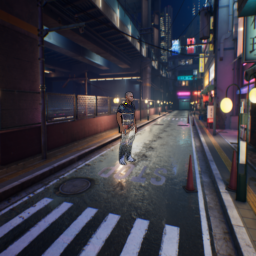}
	}
	\subfigure[Camouflage]{
		\includegraphics[height=0.145\linewidth]{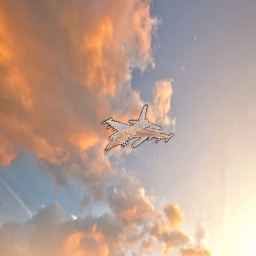}
	}
	
	\subfigure[DE$\_$DAC]{
		\includegraphics[height=0.145\linewidth]{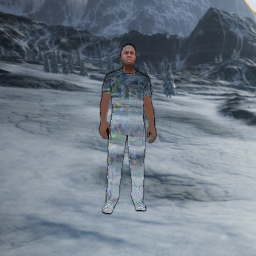}
	}
	\subfigure[DE$\_$DAC]{
		\includegraphics[height=0.145\linewidth]{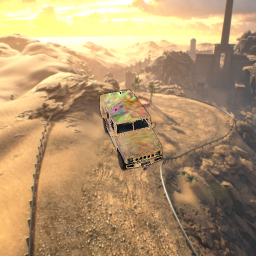}
	}
	\subfigure[DE$\_$DAC]{
		\includegraphics[height=0.145\linewidth]{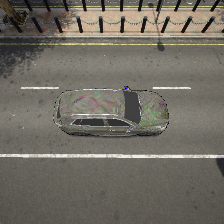}
	}	
	\subfigure[DE$\_$DAC]{
		\includegraphics[height=0.145\linewidth]{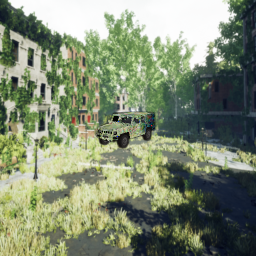}
	}
	\subfigure[DE$\_$DAC]{
		\includegraphics[height=0.145\linewidth]{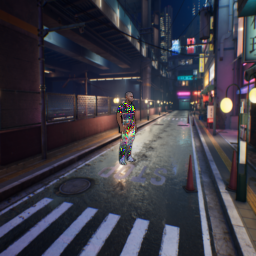}
	}
	\subfigure[DE$\_$DAC]{
		\includegraphics[height=0.145\linewidth]{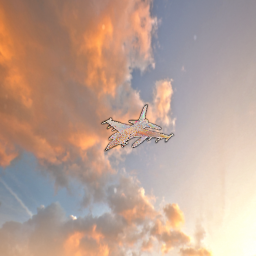}
	}
	
	\caption{The visualization of attacking other objects by our method .}
	\label{otherobject}
\end{figure}

To illustrate the generality of our proposed method, we also conduct experiments on attacking other objects, including the car, trucks and planes. The selected scenes include the city streets, sky, residential buildings and wild road. The visualization of AEs by attacking these objects is presented in Figure \ref{otherobject}. Figure \ref{otherobject} (a) to Figure \ref{otherobject} (f) are the camouflage optimized by the first stage, while Figure \ref{otherobject} (g) and Figure \ref{otherobject} (l) are the final camouflage optimized by DE$\_$DAC. All of them can cause the dropping of the accuracy of the classifier or detector models. From Figure \ref{otherobject}, we can see that the camouflage by the first stage in our proposed method can possess a superior ability to adapt well to the surrounding environment. Our proposed method can learn well camouflage to fool human eyes and object detectors not only in relatively monochrome scenes like Figure \ref{otherobject} (b) and Figure \ref{otherobject} (c) but also in complicated scenes such as Figure \ref{otherobject} (a). It means that by the neural render, our method can learn the camouflage with rich background environmental features. Besides, our proposed method can possess better flexibility in the real world. For instance, Figure \ref{otherobject} (a) and Figure \ref{otherobject} (e) present the visualization of practical camouflage in the person, which is more practical in application. In addition, our method can also be easily transferred to other objects.

\section{Conclusion and discussions}\label{sec5}
In this paper, we propose a method called differential evolution based dual adversarial camouflage to fool human eyes and object detectors simultaneously. In this method, we first propose a two-stage training strategy to train the texture that can be pasted on the surface of objects. In the first stage, we train the global texture to minimize the discrepancy between the objects and the background environments. In the second stage, we train the partial texture, denoted as the local texture, to fool the object detectors. In addition, given a 3D object and the maximum number of faces area to optimize, we introduce the DE algorithm to search for the near-optimal area. In digital-world experiments including three datasets, our proposed method can obtain the best performance among the comparison methods. Compared with pure camouflage, with a slight drop in MSE between the render object and background, our method can improve the adversarial performance, such as ASR largely, from 30$\%$ to about 80$\%$. Under the same number of faces to be optimized in the second stage, our proposed method can also improve the ASR by about 5$\%$. In addition, experiments show that our proposed end-to-end method can also be easily combined to other scenes and objects. Due to material constraints, we did not explore physical experimental effects. In future work, we would further study the performance of physical attack of DE$\_$DAC, including realizing the function of real-time colour change and camouflage.
\section*{Acknowledgments}
This work was supported by the National Natural Science Foundation of China (Nos.11725211 and 52005505).

\bibliography{mybibfile}

\end{document}